\documentclass{article}

\usepackage{microtype}
\usepackage{graphicx}
\usepackage{subcaption}
\usepackage{booktabs} 
\usepackage{hyperref}

\usepackage[preprint]{icml2026}
\usepackage{custom_header}

\usepackage{amsmath,amsfonts,bm}

\def\eqref#1{equation~\ref{#1}}

\def\1{\bm{1}}

\def\rvx{{\mathbf{x}}}
\def\rvy{{\mathbf{y}}}

\def\vy{{\bm{y}}}

\DeclareMathAlphabet{\mathsfit}{\encodingdefault}{\sfdefault}{m}{sl}
\SetMathAlphabet{\mathsfit}{bold}{\encodingdefault}{\sfdefault}{bx}{n}

\def\gD{{\mathcal{D}}}

\def\gP{{\mathcal{P}}}

\def\gS{{\mathcal{S}}}

\usepackage{amsmath}
\usepackage{amssymb}
\usepackage{mathtools}
\usepackage{amsthm}

\usepackage[most]{tcolorbox}
\usepackage{newtxmath}
\usepackage{pifont} %

\expandafter\let\csname algorithm*\endcsname\relax
\expandafter\let\csname endalgorithm*\endcsname\relax
\usepackage[ruled,vlined,linesnumbered,algo2e]{algorithm2e}
\usepackage{array}
\usepackage{pifont}%
\usepackage{inconsolata}
\usepackage{xspace}
\usepackage{enumitem}
\usepackage{wrapfig}

\tcbuselibrary{listings,skins,breakable}
\tcbuselibrary{listings}
\newcommand{\cmark}{\ding{51}}%
\newcommand{\xmark}{\ding{55}}%
\definecolor{darkmagenta}{rgb}{0.56, 0.0, 1.0} 
\hypersetup{
colorlinks,
linkcolor={blue!64!black},
citecolor={darkmagenta!72!black},
urlcolor={blue!64!black}
}

\lstdefinestyle{editor}{
  language=Python,
  basicstyle=\ttfamily\scriptsize,
  numbers=left, numberstyle=\tiny, numbersep=6pt,
  showstringspaces=false,
  columns=fullflexible,
  keepspaces=true,
  breaklines=true, breakatwhitespace=false,
  tabsize=2,
  keywordstyle=\color{blue!70!black}\bfseries,
  stringstyle=\color{green!40!black},
  commentstyle=\color{black!55},
}

\renewcommand{\cmark}{\ding{51}}
\renewcommand{\xmark}{\ding{55}}

\newtcolorbox{expblock}[2][]
{%
  breakable,
  enhanced,
  colback        = gray!2,          
  colframe       = gray!65!black,   
  coltitle       = white,           
  boxed title style = {             
    colback = gray!14,
    size    = small,
    boxrule = 0pt,                  
  },
  arc           = 1.6mm,            
  boxrule       = .4pt,             
  left          = 1.8mm,
  right         = 1.8mm,
  top           = 1.2mm,
  bottom        = 1.2mm,
  fonttitle     = \bfseries\normalsize,
  title         = {#2},
  #1
}

\usepackage{xurl}
\usepackage[capitalize,noabbrev]{cleveref}

\theoremstyle{plain}

\theoremstyle{definition}

\theoremstyle{remark}

\makeatletter

\crefname{section}{\S\@gobble}{\S\@gobble}
\crefname{subsection}{\S\@gobble}{\S\@gobble}
\crefname{proposition}{Prop.}{Props.}
\crefname{figure}{Fig.}{Figs.}
\crefname{table}{Table}{Tables}

\def\section{\@startsection{section}{1}{\z@}{-0.10in}{0.02in}
  {\large\bf\raggedright}}
\def\subsection{\@startsection{subsection}{2}{\z@}{-0.08in}{0.01in}
  {\normalsize\bf\raggedright}}
\def\subsubsection{\@startsection{subsubsection}{3}{\z@}{-0.08in}{0.01in}
  {\normalsize\sc\raggedright}}
\def\paragraph{\@startsection{paragraph}{4}{\z@}{0ex}{-1em}{\normalsize\bf}}
\def\subparagraph{\@startsection{subparagraph}{5}{\z@}{1.5ex plus
    0.5ex minus .2ex}{-1em}{\normalsize\bf}}

\makeatother

\newcommand{\Cone}[1]{\colorbox{blue!20}{#1}}   
\newcommand{\ConeM}[2][blue!20]{\mathchoice%
  {\colorbox{#1}{$\displaystyle#2$}}%
  {\colorbox{#1}{$\textstyle#2$}}%
  {\colorbox{#1}{$\scriptstyle#2$}}%
  {\colorbox{#1}{$\scriptscriptstyle#2$}}}
\newcommand{\Ctwo}[1]{\colorbox{green!20}{#1}}   
\newcommand{\CtwoM}[2][green!20]{\mathchoice%
  {\colorbox{#1}{$\displaystyle#2$}}%
  {\colorbox{#1}{$\textstyle#2$}}%
  {\colorbox{#1}{$\scriptstyle#2$}}%
  {\colorbox{#1}{$\scriptscriptstyle#2$}}}
\newcommand{\Cfour}[1]{\colorbox{orange!20}{#1}}  
\newcommand{\CfourM}[2][orange!20]{\mathchoice%
  {\colorbox{#1}{$\displaystyle#2$}}%
  {\colorbox{#1}{$\textstyle#2$}}%
  {\colorbox{#1}{$\scriptstyle#2$}}%
  {\colorbox{#1}{$\scriptscriptstyle#2$}}}
\newcommand{\NEW}[1]{\colorbox{red!20}{#1}}  

\newcommand{\std}[1]{\scriptsize$\pm$\text{#1}}

\newcommand{\method}{RSA\xspace}

\newcommand{\ie}{\textit{i.e.}}
\newcommand{\eg}{\textit{e.g.}}

\usepackage[textsize=tiny]{todonotes}

\icmltitlerunning{Recursive Self-Aggregation Unlocks Deep Thinking in Large Language Models}

\begin{document}

\onecolumn
\setheadertitle{Recursive Self-Aggregation Unlocks Deep Thinking in Large Language Models}
\setheaderauthors{%
Siddarth Venkatraman$^{*\;1,2}$
\quad 
Vineet Jain$^{*\;1,3}$
\quad
Sarthak Mittal$^{*\;1,2}$
\quad
Vedant Shah$^{1,2}$
\quad
Johan Obando-Ceron$^{1,2}$
\\\bf
Yoshua Bengio$^{1,2,4,7}$
\quad
Brian Bartoldson$^{5}$
\quad
Bhavya Kailkhura$^{5}$
\quad
Guillaume Lajoie$^{1,2,7}$
\quad
Glen Berseth$^{1,2,7}$
\\\bf
 Nikolay Malkin$^{6,8}$
\quad
 Moksh Jain$^{1,2}$
}
\setheaderaffiliations{%
\textsuperscript{1}Mila -- Qu\'ebec AI Institute
\quad\textsuperscript{2}Universit\'e de Montr\'eal
\quad\textsuperscript{3}McGill University
\quad\textsuperscript{4}LawZero
\quad\textsuperscript{5}LLNL\\
\textsuperscript{6}University of Edinburgh
\quad\textsuperscript{7}CIFAR AI Chair
\quad\textsuperscript{8}CIFAR Fellow
}
\setheadernotes{$^{*}$Equal Contribution}
\setheaderemails{%
$\left\{
    \begin{array}{@{}l@{}}
    \text{siddarth.venkatraman,jain.vineet,mittalsa,moksh.jain}
    \end{array}
  \right\}$@mila.quebec
}

\makecustomheader

\begin{abstract}
Test-time scaling methods improve the capabilities of large language models (LLMs) by increasing the amount of compute used during inference to make a prediction. Inference-time compute can be scaled \emph{in parallel} by choosing among multiple independent solutions or \emph{sequentially} through self-refinement. We propose Recursive Self-Aggregation (\method), a test-time scaling method inspired by evolutionary methods that combines the benefits of both parallel and sequential scaling. Each step of \method refines a population of candidate reasoning chains through aggregation of subsets to yield a population of improved solutions, which are then used as the candidate pool for the next iteration. Empirically, \method delivers substantial performance gains with increasing compute budgets across diverse tasks, model families and sizes. Notably, \method with Gemini 3 Flash attains performance near the top of the ARC-AGI-2 public leaderboard. \method also enables Qwen3-4B-Instruct-2507 to achieve competitive performance with larger reasoning models, including DeepSeek-R1 and o3-mini (high), outperforming purely parallel and sequential scaling strategies across AIME-25, HMMT-25, Reasoning Gym, LiveCodeBench-v6, and SuperGPQA. We further propose a novel aggregation-aware reinforcement learning approach that yields significant performance gains by training the model to combine solutions. 


\vspace{-0.75em}
\begin{center}
\href{https://github.com/rsa-llm/RSA}{%
  \raisebox{-0.15\height}{\includegraphics[height=1.2em]{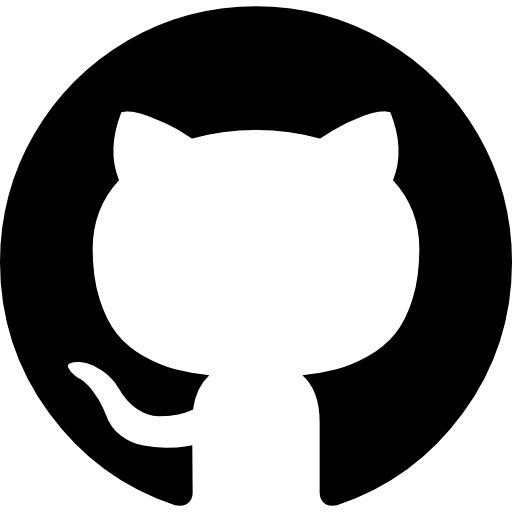}}\;%
  \texttt{rsa-llm/RSA}%
}\qquad\qquad
\href{https://github.com/rsa-llm/RSA-ARC}{%
  \raisebox{-0.15\height}{\includegraphics[height=1.2em]{logo/github-logo.png}}\;%
  \texttt{rsa-llm/RSA-ARC}%
}
\end{center}

\end{abstract}


\begin{figure}[H]
    \vspace*{-1em}
    \centering
    \includegraphics[width=\textwidth]{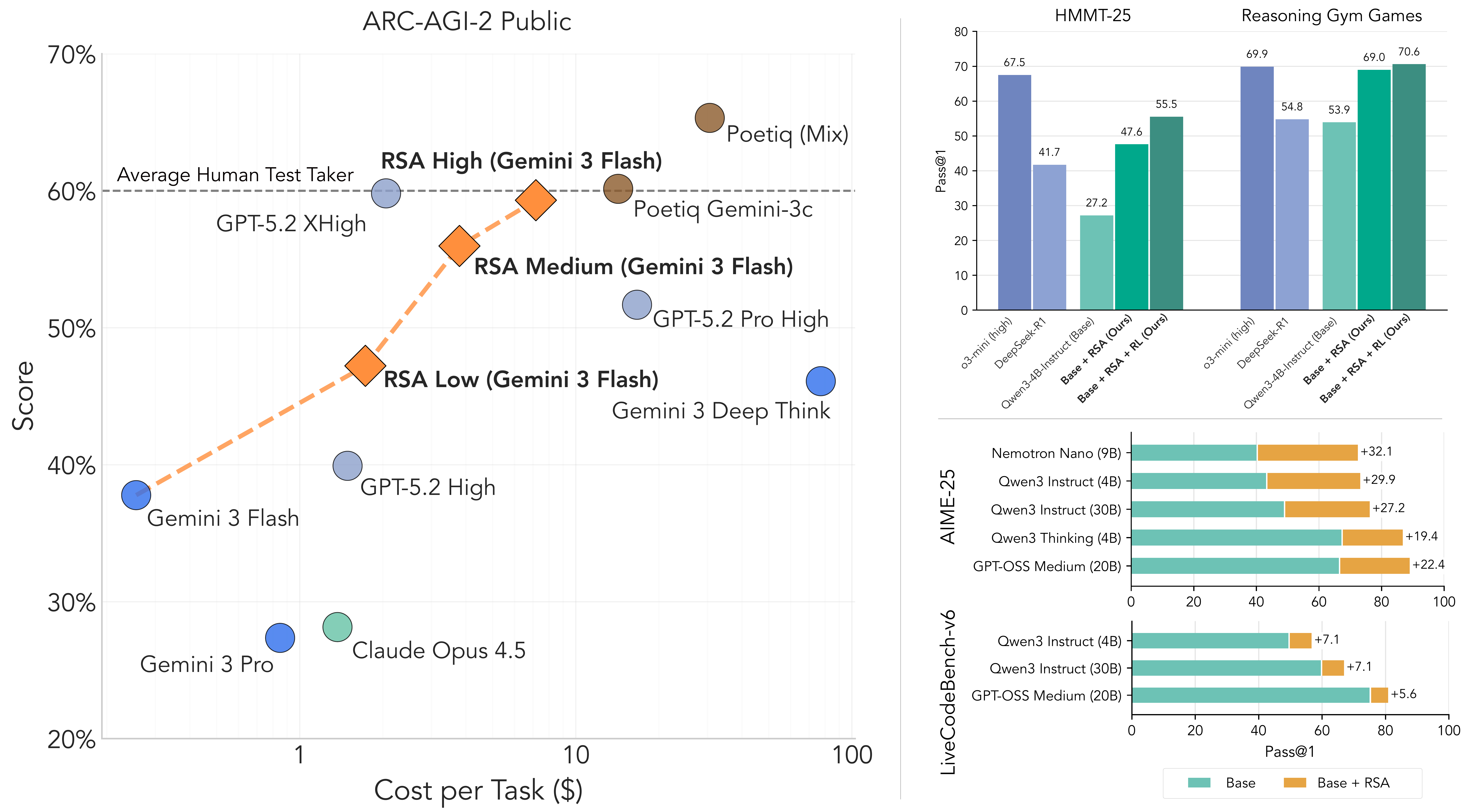}
    \caption{\textbf{Left.} \method pushes Gemini 3 Flash to near human-level performance on ARC-AGI-2, exceeding models like Gemini Deep Think at a significantly lower cost. We use RSA with $N=16, K=4$ and the Low, Medium and High configurations refer to $T=2,5,9$ respectively. \textbf{Right}. \method enables Qwen3-4B-Instruct-2507 to match the performance of larger reasoning models such as DeepSeek-R1 and o3-mini (high). 
    These gains are further amplified through our proposed aggregation-aware RL framework (\Cref{sec:rl}).}
    \label{fig:intro}
\end{figure}

\section{Introduction}
\label{sec:intro}
\looseness=-1
Large language models (LLMs) demonstrate consistent improvements in performance with increasing training compute~\citep{kaplan2020scaling}.
Complementarily, \emph{test-time scaling} strategies, \ie, those that increase compute at inference without altering model parameters, can deliver significant gains in performance~\citep{snell2025scaling,jaech2024openai}. 
Test-time scaling mechanisms for LLMs can broadly be characterised into two types (\Cref{sec:taxonomy}): those that use deeper model rollouts to iteratively improve solutions \citep[\eg,][]{muennighoff2025s1simpletesttimescaling,zhanggenerative} and those that branch to explore multiple solution paths, then filter or recombine them \citep[\eg,][]{wang2023selfconsistency, weng2023large}. We refer to these types as \emph{sequential} and \emph{parallel} scaling; some \emph{hybrid} methods combine the strengths of both approaches~\citep[\eg,][]{yao2023tree,DBLP:journals/telo/MeyersonNBGKHL24,lee2025evolvingdeeperllmthinking}.

\looseness=-1
However, a universal and effective test-time-scaling method that allows reuse of promising fragments from multiple candidate solutions is lacking. Self-refinement methods -- the quintessential form of sequential scaling -- can improve a candidate solution by reusing its own correct parts, but do not leverage the information contained within other candidates. Similarly, parallel scaling methods such as verifier-guided Best-of-N selection can identify the best candidate from a batch, but do not recombine candidates to produce improved solutions. Existing hybrid approaches fail to solve this problem in a general way, often making strong assumptions on the form of reasoning chains \citep[\eg,][]{DBLP:journals/telo/MeyersonNBGKHL24, hemberg2024evolvingcodelargelanguage} or requiring external verifiers \citep[\eg,][]{novikov2025alphaevolve, lee2025evolvingdeeperllmthinking}. Our work fills this gap in three ways, described in the following paragraphs.

\paragraph{Self-aggregation.} We study a general way to improve LLM reasoning chains through \emph{self-aggregation}: providing the model with the query and a set of candidate solutions and prompting it to produce an improved solution. Such an approach, which relies on the implicit verification abilities of the model~\citep{weng2023large}, can use the rich information contained within the reasoning chains: for example, a reasoning trace that results in an incorrect answer to a problem can have correct intermediate steps that can be reused in the aggregated solution (\Cref{app:qual_example}). Such aggregation methods are explored in multiple concurrent works \citep[\eg,][]{li2025llms, wang2025learningrefineselfrefinementparallel,zhao2025majority} and are promising directions for test-time scaling.

\paragraph{Recursive self-aggregation.}  While self-aggregation can be used as a one-time procedure to combine candidate solutions, our proposed algorithm, \emph{Recursive Self-Aggregation} (\method, \Cref{sec:rsa}), goes further by integrating aggregation steps into a self-improvement loop motivated by evolutionary algorithms. \method maintains a population of candidate solutions and iteratively recombines subsets of the population to produce a new population of improved solutions (\Cref{fig:method}). This sequential refinement enables deeper reasoning by allowing the model to revisit its solutions and make multiple attempts at correcting errors. \method maintains a candidate population larger than the aggregation set size and can therefore jointly consider significantly more proposals than single-step aggregation, which is constrained by the model’s effective context length. Unlike other evolutionary methods, \method requires no external verification and can be seamlessly integrated into any LLM inference pipeline to improve reasoning.

\looseness=-1
\paragraph{Aggregation-aware RL.} During post-training, LLMs are trained with reinforcement learning (RL) to improve their reasoning ability~\citep{jaech2024openai,deepseekai2025deepseekr1incentivizingreasoningcapability}. RL training does not account for the test-time scaling that will be applied to the post-trained model. In fact, we observe that standard RL fine-tuning can even degrade performance relative to the base model when combined with test-time aggregation (\Cref{sec:rl_exp}). To address this, we propose an \emph{aggregation-aware} RL approach, a simple data-augmentation strategy to train LLMs to aggregate solutions (\Cref{sec:rl}).

\method pushes Gemini 3 Flash to near the top of the ARC-AGI-2 public leaderboard (\Cref{sec:arc_agi}). We also perform extensive experiments with open models to demonstrate the effectiveness of \method across diverse tasks, such as AIME-25, HMMT-25, LiveCodeBench, Reasoning Gym, and SuperGPQA with various base models (\Cref{sec:exp}). \method bridges the gap between the lightweight \texttt{Qwen3-4B-Instruct-2507} and much larger reasoning models like \texttt{DeepSeek-R1} and \texttt{o3-mini (high)} (\cref{fig:intro}). Our results also show that aggregation-aware RL significantly improves performance with \method compared to naïve RL training (\cref{sec:rl_exp}). We rigorously analyze the factors driving \method performance and provide practical recommendations to enable deeper test-time thinking under compute constraints (\Cref{sec:ablations}).

\section{A Taxonomy of Test-Time Scaling Methods}
\label{sec:taxonomy}

\begin{figure}[t]
    \centering
    \includegraphics[width=\linewidth]{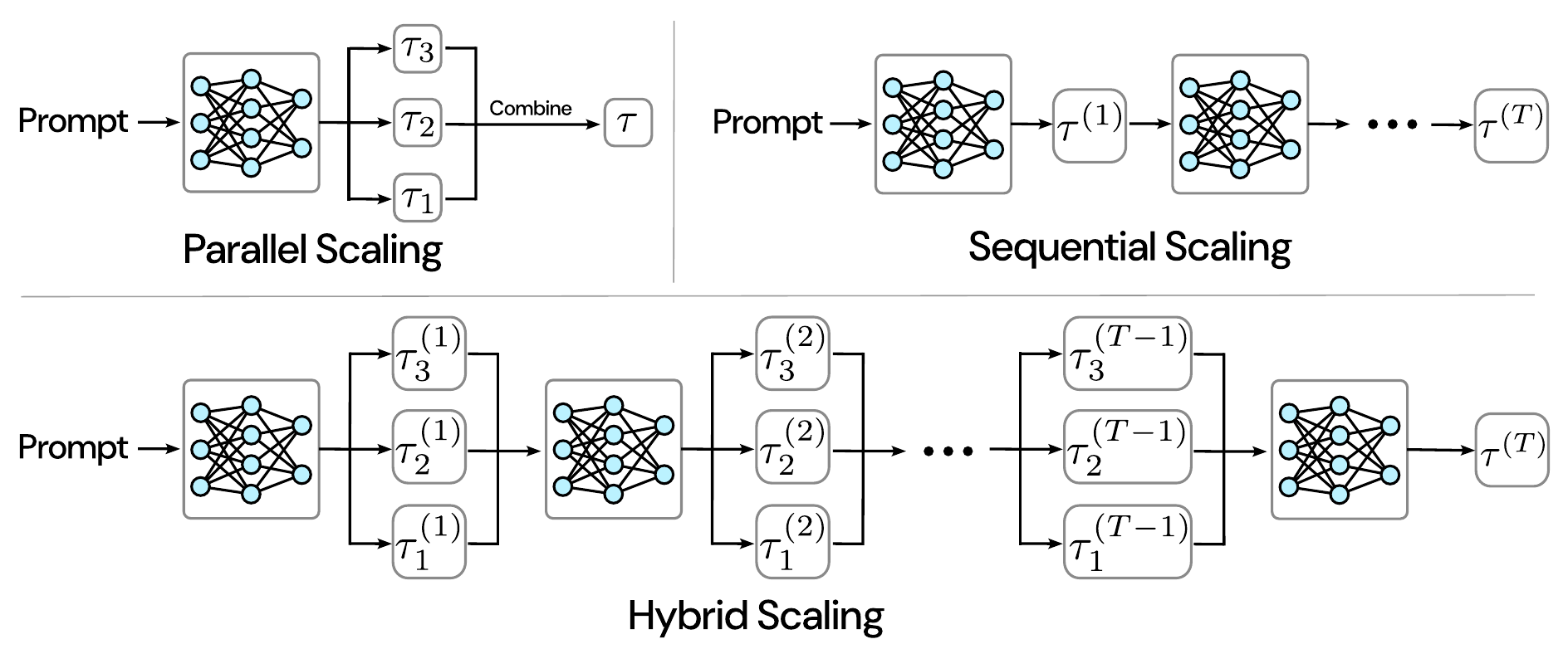}
    \caption{\textbf{Overview of test-time scaling control flows.} \emph{Parallel} methods generate multiple candidates and select using a verification mechanism. \emph{Sequential} methods iteratively refines a chain, correcting previous mistakes. \emph{Hybrid} methods combine parallel branching with sequential refinement.}
    \label{fig:tts_diagram}
    \vspace*{-1.5em}
\end{figure}

\emph{Test-time scaling} refers to methods that obtain predictions using a static LLM with a larger number of model evaluations than that required by simply prompting for an answer. These methods significantly improve performance without modifying model weights~\citep{snell2025scaling,zhang2025surveytesttimescalinglarge}, effectively using the model as a component in an external optimization or inference framework, at the cost of increased computation.

\looseness=-1
A well-designed test-time scaling framework should yield monotonic improvements in performance as compute budgets increase, similar to scaling laws for pretraining~\citep{kaplan2020scaling,snell2025scaling}.
Most methods rely on some kind of verification, whether implicit or explicit, incorporated within a sequential or parallel control flow.
In this section, we review the literature on test-time scaling in LLMs and provide a taxonomy of test-time scaling frameworks based on the verification strategy and control flow they employ, illustrated in \cref{fig:tts_diagram}. Building on this, we then introduce our proposed approach, \textit{Recursive Self-Aggregation} (\method) in \Cref{sec:rsa}. 
See \Cref{app:related_work} for a discussion of broader related work.

\subsection{Verification Strategy}

\looseness=-1
\paragraph{External verification.} Any external optimization procedure requires a mechanism to assess the quality of proposed solutions. In domains such as code or math, verification can often be performed exactly using external tools (\eg, compilation and execution \citep{gao2023pal}). If exact verifiers are unavailable, inference-time feedback is instead obtained via \emph{learned} reward models, trained on preference data or correctness signals derived from reasoning chains~\citep{cobbe2021training,ouyang2022traininglanguagemodelsfollow,snell2025scaling}.
Verifier feedback, exact or learned, makes it possible to improve solution quality as more compute is allocated: a simple strategy is Best-of-N~\citep{gao2023scaling}, where the highest-reward solution out of $N$ generated candidates is selected. 

\paragraph{Self-verification.} LLMs exhibit a generation-verification gap: they are more reliable at judging the correctness of solutions than producing them \citep{li2024benchmarking}. This property can be exploited to enable test-time scaling by using the LLM as a verifier of its own outputs \citep[\eg,][]{madaan2023selfrefine, weng2023large}. The LLM can also be fine-tuned to enhance its verification ability \citep{zhanggenerative}, but we consider this a form of external verification since it requires learning a verifier.

\paragraph{Implicit verification.} Some methods bypass explicit verification by relying on the LLM to generate improved solutions, effectively performing verification of solutions without scoring them. For example, majority voting \citep{wang2023selfconsistency} works on the assumption of self-consistency: that the model produces correct answers more consistently than incorrect ones. Similarly, self-refinement frameworks \citep[\eg,][]{madaan2023selfrefine} iteratively refine a reasoning chain without being explicitly prompted for verification.
\method falls within this category: rather than explicitly verifying each solution, the model implicitly checks intermediate steps across multiple reasoning chains, allowing it to correct errors and generate improved solutions.

\subsection{Reasoning Control Flow}

\paragraph{Parallel scaling.} These strategies generate multiple independent reasoning chains in parallel and then combine them to yield the final answer. Typical procedures for combination include majority voting, Best-of-N selection or single-step aggregation of the sampled proposals~\citep{wang2023selfconsistency,snell2025scaling,li2025llms}. Such methods rely on the inherent diversity in sampling from the LLM, allowing parallel proposals to efficiently explore the search space and allow optimal GPU memory utilization. These algorithms embody the philosophy of \emph{breadth-first thinking}.

\paragraph{Sequential scaling.} Purely parallel scaling sacrifices the ability to think deeply, which is often crucial for multi-step reasoning tasks that cannot be solved efficiently through guess-and-check. 
Sequential scaling instead increases the number of iterative model evaluations to produce higher-quality solutions, for example, by inducing a model to correct errors in its reasoning \citep{muennighoff2025s1simpletesttimescaling} or simply increasing the number of latent reasoning tokens it generates.
While these strategies generally require more computation time than parallel ones (given sufficient memory budgets), they are well suited to complex reasoning problems requiring \emph{depth-first thinking}. 
However, the lack of branching in such methods limits their ability to explore alternative continuations of promising solution paths, making the model prone to persisting in an unproductive reasoning chain. Sequential scaling also leaves excess GPU memory underutilized.

\paragraph{Hybrid scaling.} Sequential and parallel scaling strategies can be combined in hybrid frameworks that draw on the strengths of both paradigms. These methods make efficient use of GPU memory by evaluating many candidate solutions in parallel, while also incorporating sequential depth to iteratively refine and improve the batch of solutions. One strong class of hybrid methods uses LLMs as components within a genetic algorithm loop \citep[\eg,][]{DBLP:journals/telo/MeyersonNBGKHL24,novikov2025alphaevolve, lee2025evolvingdeeperllmthinking}, all using external verification to score candidates. Another example of hybrid test-time scaling is Mixture-of-Agents \citep{wang2024mixture}, where an ensemble of LLMs generates improved proposals that are aggregated by a strong model into the seed solution for the next iteration. Our method, \method, is also a hybrid scaling algorithm: like Mixture-of-Agents, it relies on recursive aggregation, but it further maintains a population of candidate solutions larger than the aggregation batch size, similar to evolutionary algorithms, while only using a single LLM. By aggregating random subsets of this population, \method preserves diversity in the candidate pool, which is critical when all proposals and aggregations are produced by the same model (as studied in \Cref{sec:n_ablation,app:diversity_analysis}).

\section{Evolving Thoughts using Recursive Self-Aggregation}
\label{sec:rsa}
\begin{figure*}[t]
    \centering
    \includegraphics[width=\linewidth]{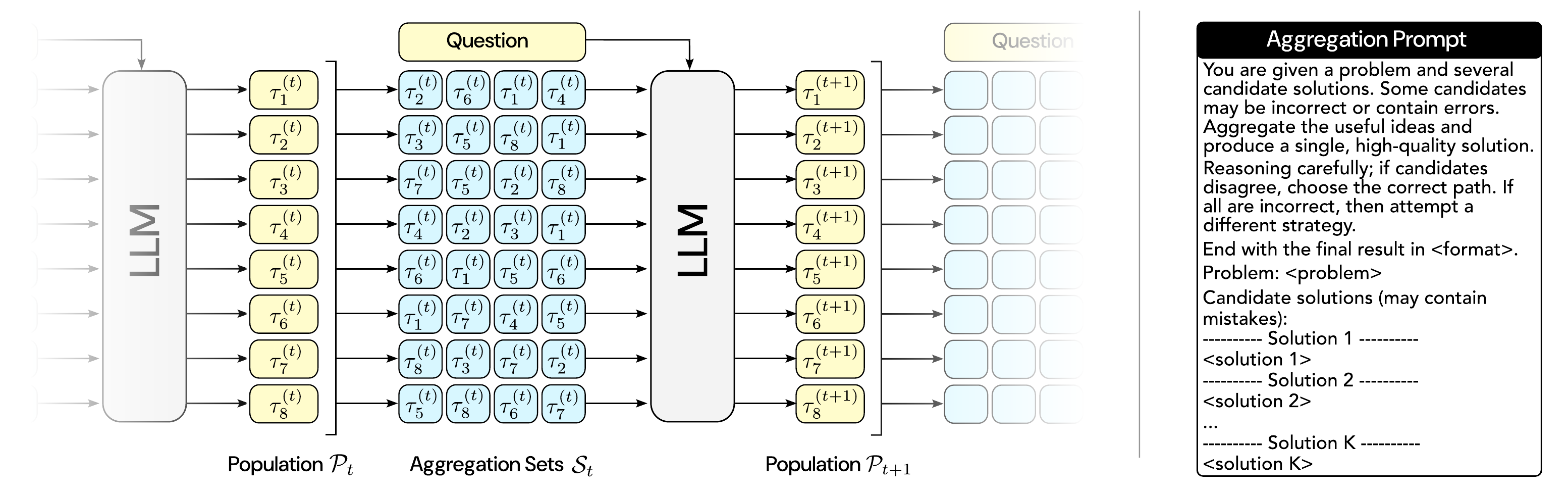}
    \caption{\method generates a population of $N$ solutions for a given prompt and recursively updates them over $T$ steps. Each update step subsamples $K$ distinct solutions from the current population and generates an improved solution with the aggregation prompt. See \Cref{app:algorithm} for algorithm pseudo-code.}
    \label{fig:method}
    \vspace*{-1em}
\end{figure*}

We present \textit{Recursive Self-Aggregation} (\method), a hybrid test-time scaling procedure designed to improve the model's performance without complex scaffolding or using external verifiers. It frames reasoning as a form of evolutionary process where candidate reasoning chains are iteratively refined through self-aggregation, inspired by the crossover and mutation steps in genetic algorithms. 
\method is simple to implement and leads to substantial improvements in reasoning abilities across different models and tasks, when compared to pure sequential or parallel scaling (\Cref{sec:exp}). \Cref{fig:method} illustrates the core components of \method, which we describe below. The algorithm is also written in \Cref{app:algorithm}.

Given a query $\rvx$ and a pretrained LLM $p_{\theta_\textrm{ref}}$, \method maintains a population of $N$ candidate solutions $\gP_t$ at each step $t$. The model is provided with the question and a subset of $K$ solutions from this population, and prompted to produce an improved population of solutions $\gP_{t+1}$. The procedure is described in detail below:

\begin{enumerate}[left=0cm,nosep]
    \item \textbf{Population of trajectories.} At any given step $t$, \method maintains a population of $N$ independent candidate solutions $\gP_t := \{\tau_{1}^{(t)}, \dots \tau_{N}^{(t)}\}$. The initial population $\gP_1$ is generated by sampling $N$ responses for query $\rvx$ using the LLM $p_{\theta_\textrm{ref}}$:
    \begin{equation}
    \tau_i^{(1)} \sim p_{\theta_\textrm{ref}}(\,\cdot \mid \rvx), \quad \mathcal{P}_{1} = \{\tau_1^{(1)}, \dots, \tau_N^{(1)}\}.
    \end{equation}
    \item \textbf{Subsampling}. We form $N$ aggregation sets of $K$ candidates, where each set is sampled uniformly without replacement from the population:
    \begin{equation}
    \label{eq:subsampling}
    \mathcal{S}_t = \{ S_1^{(t)}, S_2^{(t)}, \dots, S_N^{(t)} \}, \ S_i^{(t)} \subseteq \mathcal{P}_t, \ |S_i^{(t)}| = K.
    \end{equation}
    \item \textbf{Aggregation.} Each set $\gS_i^{(t)}$ and the query $\rvx$ is formatted using an aggregation prompt directing the LLM $p_{\theta_\textrm{ref}}$ to generate a refined response $\tau_{i}^{(t+1)}$, forming a new population of candidates $\gP_{t+1}$:
    \begin{equation}
    \label{eq:agg}
    \tau_i^{(t+1)} \sim p_{\theta_\textrm{ref}}(\,\cdot \mid S_i^{(t)},\rvx), \quad \mathcal{P}_{t+1} = \{\tau_1^{(t+1)}, \dots, \tau_N^{(t+1)}\}.
    \end{equation}
    \method recursively updates the population $\gP_t$ using \Cref{eq:subsampling} and \Cref{eq:agg} for $t=1,\dots,T-1$. This sequential loop is expected to allow errors and inconsistencies to be gradually pruned away during aggregation, while preserving favorable reasoning patterns. Consequently, we expect overall diversity within the population to generally decrease as $t$ increases, accompanied by a monotonic improvement in success rate (See \Cref{app:diversity_analysis}).
    \item \textbf{Termination.} Given the final population of candidate solutions $\gP_T$, the solution is obtained either by randomly sampling from this population or by majority voting. We use uniform random sampling in all our experiments, to evaluate our method without any special selection mechanism.
\end{enumerate}

Note that the intermediate trajectories $\tau_i^{(t)}$ are not required to terminate with complete answers; even partial reasoning chains can provide valuable signal during aggregation. Additionally, the choice of $K$ defines the number of alternative responses to consider for aggregation, with $K=1$ being equivalent to sequential self-refinement \citep{madaan2023selfrefine}. In \Cref{sec:k_ablation}, we show that even setting $K=2$ leads to significant improvements over self-refinement, highlighting the importance of combining diverse solutions for improving reasoning performance. See \Cref{app:qual_example} for an illustrative example of aggregation.

An important consideration is that self-aggregation can lead to loss of diversity due to excessive reuse of reasoning patterns that occur in multiple trajectories in the population. Maintaining a large population size $N$ relative to the aggregation size $K$ helps ensure sufficient diversity for recombination. However, a very large $N$ relative to $K$ can lead to slow convergence of the population as a whole, since high-quality reasoning patterns will require more iterations to dominate the population. We study these tradeoffs in \Cref{sec:ablations}. The aggregation prompts we use are provided in \Cref{app:prompts}.

\section{Training Aggregators with Reinforcement Learning}
\label{sec:rl}

In addition to the test-time strategies discussed thus far, a model's reasoning ability can be improved by post-training it with reinforcement learning (RL)~\citep{jaech2024openai,deepseekai2025deepseekr1incentivizingreasoningcapability}. Standard RL post-training encourages a model to produce correct solutions, conditioned on the query~\citep{luong2024reft,lambert2025tulu}. While this improves the model’s ability to directly generate correct solutions, it does not explicitly teach the model how to aggregate multiple candidate solutions. As we show in \Cref{sec:rl_exp}, this mismatch between the training objective and the test-time strategy can result in \emph{worse} performance compared to the base (reference) model when using \method.

To better align training and inference, we formulate the task of aggregation as an RL problem. The reference model $p_{\theta_\textrm{ref}}$ generates a set of candidate reasoning chains given a problem. Next, the model is trained to produce a single correct reasoning chain given the problem and the set of candidate reasoning chains. To achieve this in practice, we create an \emph{aggregation-aware} training dataset consisting of two types of prompts: (1) Standard prompts, containing only the problem, to train the model to propose good initial candidate reasoning chains; and (2) aggregation prompts, which include the problem along with $K$ candidate solutions from the reference model, formatted with the same aggregation prompt used for \method; see \Cref{app:prompts}. 

Consider problem-solution pairs sampled from some dataset $(\rvx, \rvy) \sim \gD$, and candidate solutions $\tau$ generated by the model conditioned on the problems. Training with the standard prompts described above corresponds to the standard RL training of LLMs that optimizes the following objective:
\begin{equation}
\label{eq:standard_rl}
    \max_\theta \mathbb{E}_{(\rvx, \rvy) \sim \gD}\bigl[\mathbb{E}_{\tau \sim p_\theta(\cdot \mid \rvx)}\left[r(\tau, \rvy)\right] - \beta \,\text{KL}\left(p_\theta(\cdot \mid \rvx) \,\|\, p_{\theta_\textrm{ref}}(\cdot \mid \rvx)\right)\bigr],
\end{equation}

where $\beta$ controls the optional KL regularization with the reference policy $p_{\theta_\textrm{ref}}$. For the aggregation prompts, we additionally sample $K$ candidates from $p_{\theta_\textrm{ref}}$ to construct the aggregation set $S_0$, resulting in the following objective:
\begin{equation}
\label{eq:agg_rl}
    \max_\theta \mathbb{E}_{(\rvx, \vy) \sim \gD, S_0 \sim p_{\theta_\textrm{ref}}(\cdot \mid \rvx)}\bigl[\mathbb{E}_{\tau \sim p_\theta(\cdot \mid \rvx, S_0)} \left[r(\tau, \vy) \right] - \beta \,\text{KL}\left(p_\theta(\cdot \mid \rvx, S_0) \,\|\, p_{\theta_\textrm{ref}}(\cdot \mid \rvx, S_0)\right)\bigr].
\end{equation}

This objective can be optimized using any off-the-shelf policy gradient algorithm, such as PPO~\citep{ouyang2022traininglanguagemodelsfollow}, GRPO~\citep{shao2024deepseekmathpushinglimitsmathematical}, or RLOO~\citep{ahmadian2024basicsrevisitingreinforcestyle}, initializing $\theta$ with a copy of the base model parameters $\theta_{\rm ref}$. We use RLOO in all our experiments (\Cref{sec:rl_exp}) for its simplicity and good empirical performance.

\section{Experiments}
\label{sec:exp}

We first demonstrate the effectiveness of \method as a test-time scaling strategy by evaluating it on the public evaluation set of the challenging ARC-AGI-2 benchmark \citep{chollet2025arc}. We then comprehensively evaluate \method on math, code generation, general reasoning, and knowledge recall benchmarks in \Cref{sec:test_time_comparison} and \Cref{sec:model_comparison}. \Cref{sec:ablations} analyzes how \method's key parameters -- the aggregation set size $K$, the population size $N$, and the number of sequential steps $T$ -- contribute to its success. Finally, in \Cref{sec:rl_exp}, we show that aggregation-aware RL training can further enhance \method's performance.

\paragraph{Tasks.} For open-weight models, we consider four benchmark categories, with further details in \Cref{app:tasks}:
\begin{itemize}[left=0cm,nosep]
    \item \textbf{Math.} We use AIME-25 and HMMT-25 from MathArena \citep{balunovic_srimatharena_2025}, each containing $30$ challenging competition-level math problems.
    \item \textbf{General reasoning.} We construct two datasets with $100$ problems each from Reasoning Gym \citep[RG,][]{stojanovski2025reasoninggymreasoningenvironments}, using tasks from the games and the cognition + ARC categories. The RG ARC is similar, though considerably simpler, to \citet{chollet2025arc}.
    %
    \item \textbf{Code generation.} We use LiveCodeBench-v6  \citep{jain2024livecodebenchholisticcontaminationfree} which contains $1055$ problems.  
    \item \textbf{Knowledge-based reasoning.} We use SuperGPQA \citep{pteam2025supergpqascalingllmevaluation}, a graduate-level knowledge-based reasoning benchmark, to test effectiveness of \method on tasks requiring factual recall. 
    Given its large size, we only evaluate on $1000$ randomly chosen questions. 
\end{itemize}

In addition to these tasks, we also evaluate \method on the extremely challenging ARC-AGI-2 benchmark \citep{chollet2025arc}, which tests the efficiency and capability of reasoning systems on tasks like symbolic interpretation, compositional reasoning, and contextual rule application. 

\begin{wrapfigure}{r}{0.45\textwidth}
\vspace*{-4.0em}
    \centering
    \includegraphics[width=\linewidth]{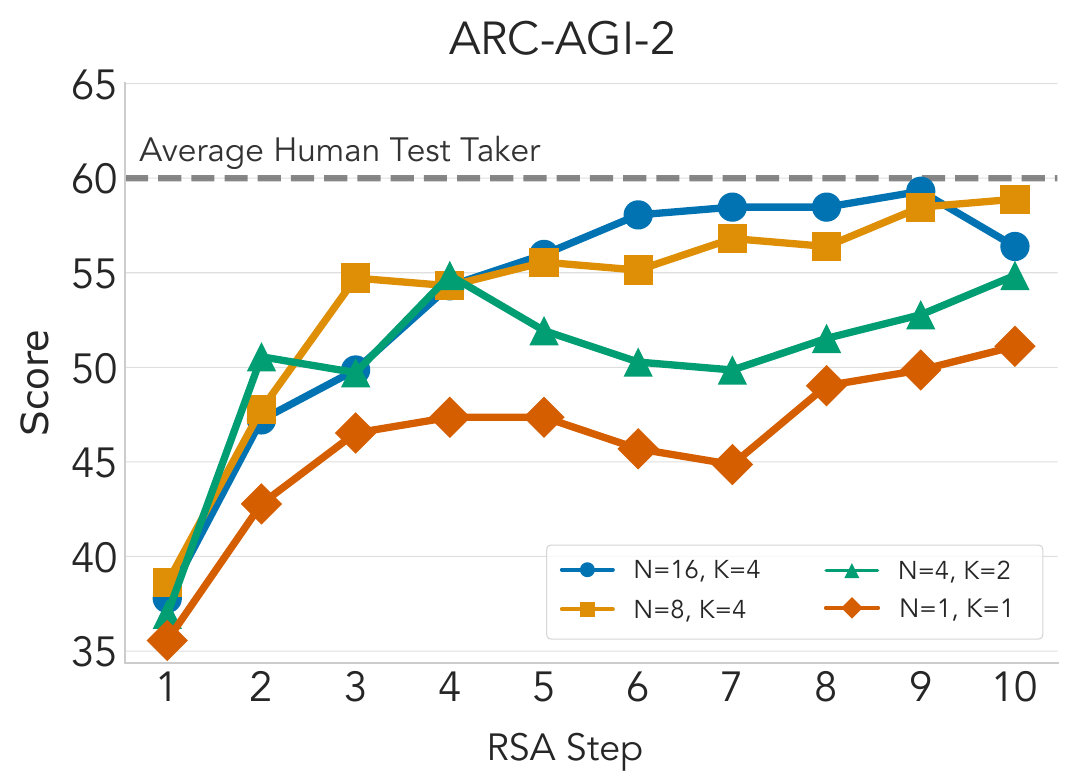}
    \vspace{-1.5em}
    \caption{ARC-AGI-2 scores for  Gemini 3 Flash + \method with different configurations of $N$ and $K$, run for $T=10$ steps.}
    \vspace{-1.5em}
    \label{fig:arc_agi}
\end{wrapfigure}

\subsection{\method Matches Human Performance on ARC-AGI-2}
\label{sec:arc_agi}

We benchmark \method with Gemini 3 Flash\footnote{via the \href{https://docs.cloud.google.com/vertex-ai/generative-ai/docs/models/gemini/3-flash}{Gemini public API}} on the ARC-AGI-2 public evaluation dataset. As the true chain-of-thought for the model is inaccessible, we instead use the summarized reasoning chains provided by the API. \method pushes the score up from $37.78\%$ (base model) to $59.31\%$ using $N=16, K=4$ and $T=10$ aggregation steps. \Cref{fig:intro} shows the cost v/s performance trade-off of different frontier models. Notably, \method with Gemini 3 Flash outperforms Gemini 3 Deep Think at $\sim10\%$ of the cost. \Cref{fig:arc_agi} shows the performance versus sequential depth for different combinations of $N$ and $K$, along with a sequential refinement baseline ($N=1, K=1$). Further details provided in \Cref{app:arc}.

\subsection{\method Outperforms Other Test-Time Scaling Methods}
\label{sec:test_time_comparison}

\begin{table*}[t]
\centering
\caption{We report Pass@1 scores for \method and other test-time scaling baselines. \method results obtained with aggregation size $K=4$, population size $N=16$, run for $T=10$ steps. Majority-voting and rejection-sampling are budget-matched with \method. Results are averaged over 4 seeds for all tasks except SuperGPQA, where we use 1 seed. Further details in \Cref{app:details}.}
\resizebox{\linewidth}{!}{
\begin{tabular}{@{}lcccccc}
\toprule
& \multicolumn{2}{c}{Math reasoning}
& Code gen.
& \multicolumn{2}{c}{General reasoning / planning}
& Knowledge recall
\\\cmidrule(lr){2-3}\cmidrule(lr){4-4}\cmidrule(lr){5-6}\cmidrule(lr){7-7}
\textbf{Method $\downarrow$ Task $\rightarrow$} & \textbf{AIME-25} & \textbf{HMMT-25} & \textbf{LiveCodeBench-v6} & \textbf{RG Games} & \textbf{RG Cognition+ARC} & \textbf{SuperGPQA}\\
\midrule
  Reference & 43.91 & 27.17 & 49.63 & 53.88 & 42.31 & 41.85 \\
  Rejection sampling  & 48.23 & 32.76 & 50.90 & 56.41 & 45.11 & 46.18 \\
  Self-refinement  & 53.33 & 39.17 & 51.40 & 65.50 & 49.17 & 43.5 \\
  Majority voting  & 68.33 & 35.00 & -- & 65.15 & 46.07 & \textbf{48.2} \\
  \midrule
  Self-aggregation (RSA, $T=1$) & 56.51\std{3.06} & 36.15\std{1.26} & 51.94\std{0.15} & 65.21\std{0.98} & 48.63\std{0.81} & 45.91\\
  \textbf{RSA}  & \textbf{73.18}\std{2.20} & \textbf{47.55}\std{1.00} & \textbf{56.72}\std{0.65} & \textbf{68.98}\std{1.10} & \textbf{51.96}\std{2.24} & 47.39\\
\bottomrule
\end{tabular}}
\label{tab:main_results}
\end{table*}

\begin{figure*}[t]
    \centering
    \includegraphics[width=\linewidth]{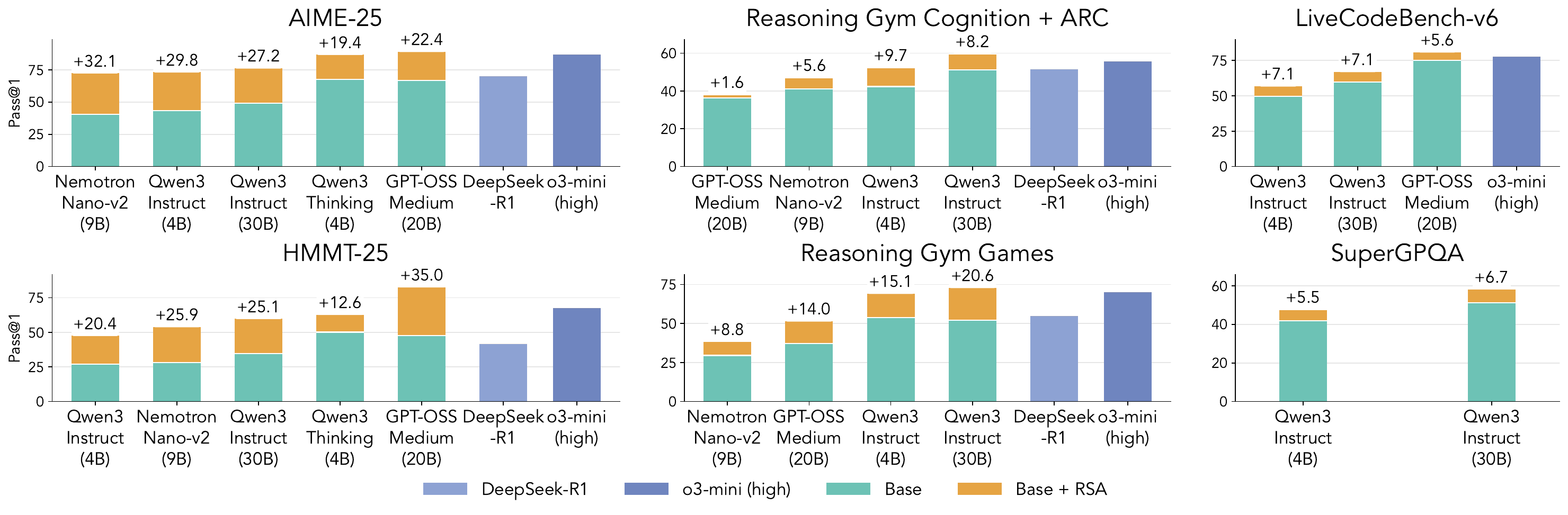}
    \caption{\textbf{\method significantly improves Pass@$1$ across math, code, general reasoning, and knowledge recall tasks.} We observe consistent gains across diverse model families, including standard instruction-tuned models and long CoT ``thinking'' models. Further details provided in \Cref{app:models_used}.}
    \label{fig:main_results_models}
    \vspace*{-1.5em}
\end{figure*}

We benchmark \method against sequential and parallel test-time scaling methods with \texttt{Qwen3-4B-Instruct-2507} as the base model. To ensure consistency across tasks, we fix the population size to $N=16$, the aggregation set size to $K=4$, and the number of recursive updates to $T=10$. Results are averaged over $4$ seeds, except for SuperGPQA where we report a single seed due to computational constraints. For fairness, we restrict comparisons to methods that require no additional training or external verifiers. Further experimental details are provided in \Cref{app:baselines}.

\begin{figure*}
    \centering
    \includegraphics[width=\linewidth]{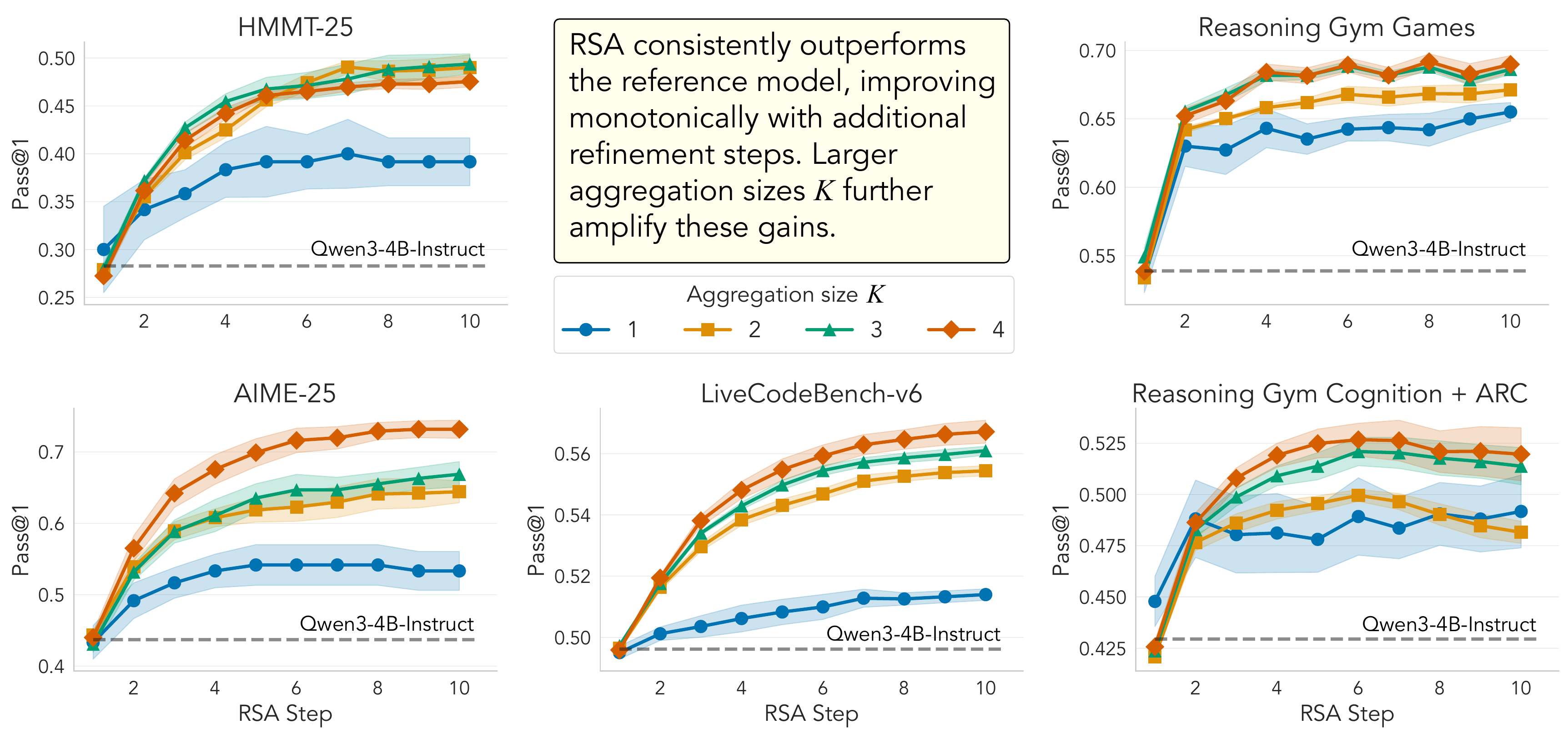}
    \caption{Pass@$1$ vs.\ \method steps, for fixed population size $N=16$, using \texttt{Qwen3-4B-Instruct-2507}. Error bands indicate standard deviation over 4 seeds. Larger $K$ generally improves performance.}
    \label{fig:acc_vs_loop}
    \vspace*{-1em}
\end{figure*}

\paragraph{Sequential baselines.} We consider $T$-step self-refinement \citep{madaan2023selfrefine}, which corresponds to \method with $K=1$ and $N=1$, which we run for $T=10$ steps.

\paragraph{Parallel baselines.} We evaluate majority voting \citep{wang2023selfconsistency} and rejection sampling with self-verification \citep{weng2023large}, budget-matched with \method through $N \times T$ generations. We also include single-step self-aggregation \citep{li2025llms}, equivalent to \method with $K=4$ and $T=1$.

\Cref{tab:main_results} reports Pass@1 results across all benchmarks, showing that \method consistently outperforms both sequential and parallel baselines. Against self-refinement, \method achieves higher performance, demonstrating that aggregating multiple solutions provides clear advantages over refining a single one. Notably, \method with $T=10$ outperforms its single-step variant ($T=1$), highlighting the benefits of recursive aggregation. When compared to parallel methods, \method achieves superior results on all tasks except SuperGPQA, where majority voting is particularly effective due to the multiple-choice answer format.
We omit majority voting on LiveCodeBench-v6 since code solutions rarely coincide exactly, and refer to \Cref{app:details} for further details.

\subsection{\method Yields Consistent Gains across Models}
\label{sec:model_comparison}
We apply \method to a diverse set of instruction-tuned models spanning various parameter counts, architectures, and reasoning abilities, including thinking models, sparse Mixture-of-Experts (MoE), and hybrid state-space models. 
\Cref{tab:model_diversity} provides an overview of the models we study. \Cref{fig:main_results_models} shows that \method leads to substantial improvements on all tasks. Remarkably, applying \method to \texttt{Qwen3-4B-Instruct-2507}, a substantially weaker model, matches and in some cases outperforms stronger models like \texttt{DeepSeek-R1} and \texttt{o3-mini (high)} without \method.
These results establish \method as a \emph{strong and general} test-time scaling strategy.

\subsection{Effect of \method Hyperparameters}
\label{sec:ablations}

We perform experiments using \texttt{Qwen3-4B-Instruct-2507} to answer the following questions:
\begin{itemize}[left=0pt,nosep]
    \item How does the performance of \method vary with parallel and sequential scaling parameters?
    \item What underlying mechanisms explain the performance gains of \method?
    \item How to select the parameters under a compute budget?
\end{itemize}

\paragraph{Monotonic improvement with sequential depth $T$.}
\label{sec:t_ablation}
\Cref{fig:acc_vs_loop} plots the Pass@1 scores over self-aggregation steps for different aggregation set sizes $K$. 
Performance improves monotonically on nearly all tasks, with the only significant downward trend on Reasoning Gym Cognition + ARC after five steps. Overall, these results demonstrate that \method scales effectively with increasing depth.

\paragraph{Increasing aggregation size $K$ improves performance.}
\label{sec:k_ablation}
\Cref{fig:acc_vs_loop} further shows improved performance with increasing aggregation size $K$. The largest gain is observed from $K=1$ to $K=2$, highlighting that aggregating over multiple reasoning chains provides substantial improvement over single-trajectory refinement. We observe diminishing returns beyond $K=3$ on most tasks, possibly because the model cannot effectively attend to very long contexts. 

\begin{wrapfigure}{r}{0.32\textwidth}
    \centering    
    \includegraphics[width=\linewidth]{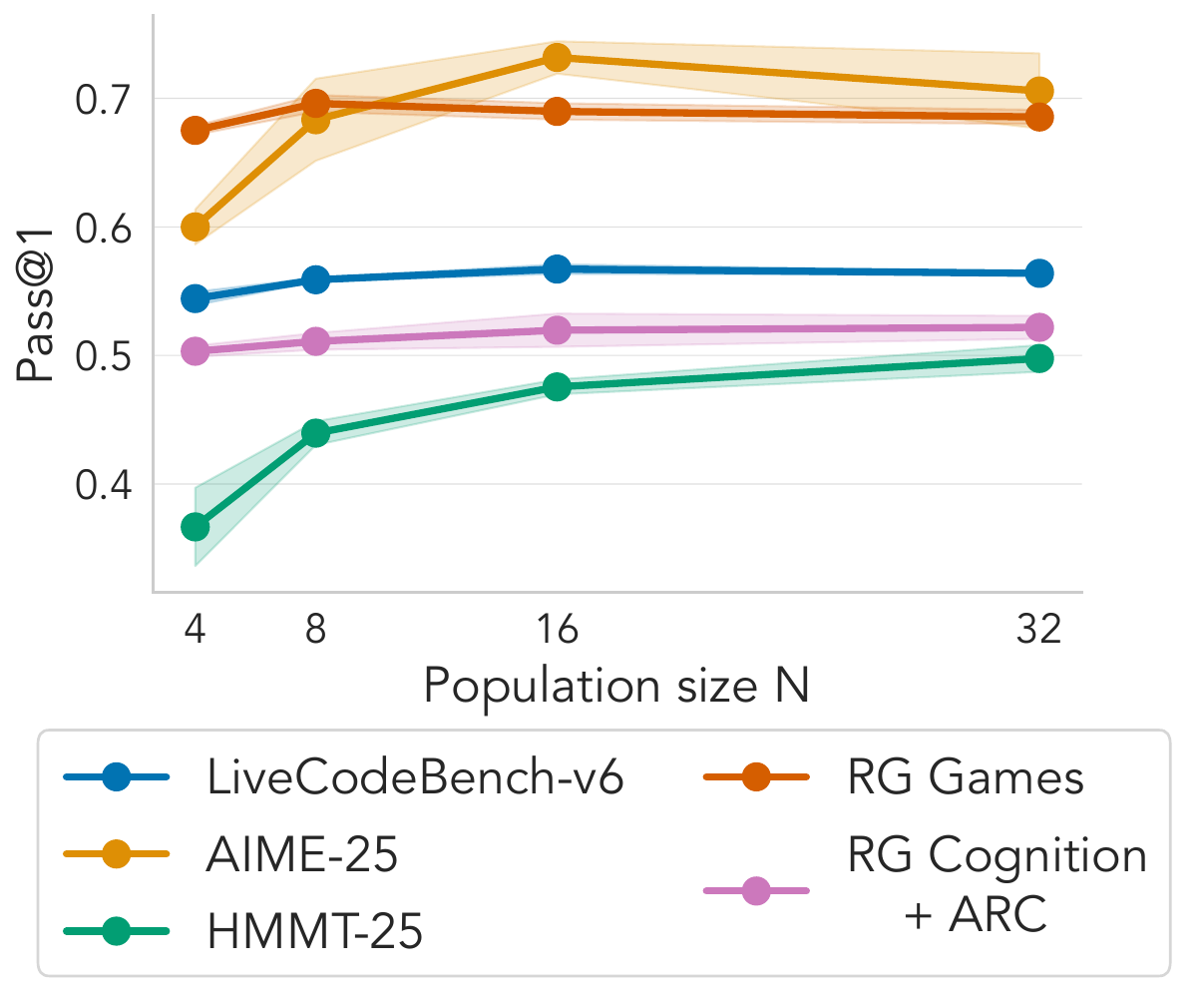}
    \caption{Pass@$1$ at $T=10$ over population size $N$ (fixed $K=4$).}
    \label{fig:n_tasks}
    \vspace*{-1em}
\end{wrapfigure}
\paragraph{Effect of increasing population size $N$.}
\label{sec:n_ablation}
Next, we study the impact of the number of candidates $N$ available for aggregation at each step.
\Cref{fig:n_tasks} shows the final Pass@$1$ scores for different tasks using $K=4$ aggregation size, $T=10$ sequential steps, and varying $N \in \{4,8,16,32\}$. 
We observe that increasing $N$ always initially improves performance, but scaling $N$ to very large values leads to a small performance drop on AIME-25 and RG Games. We investigate the role of population size further in the following section, where it emerges as the key parameter controlling the asymptotic performance of \method. 

\begin{figure*}[t]
    \centering
    \includegraphics[width=\linewidth]{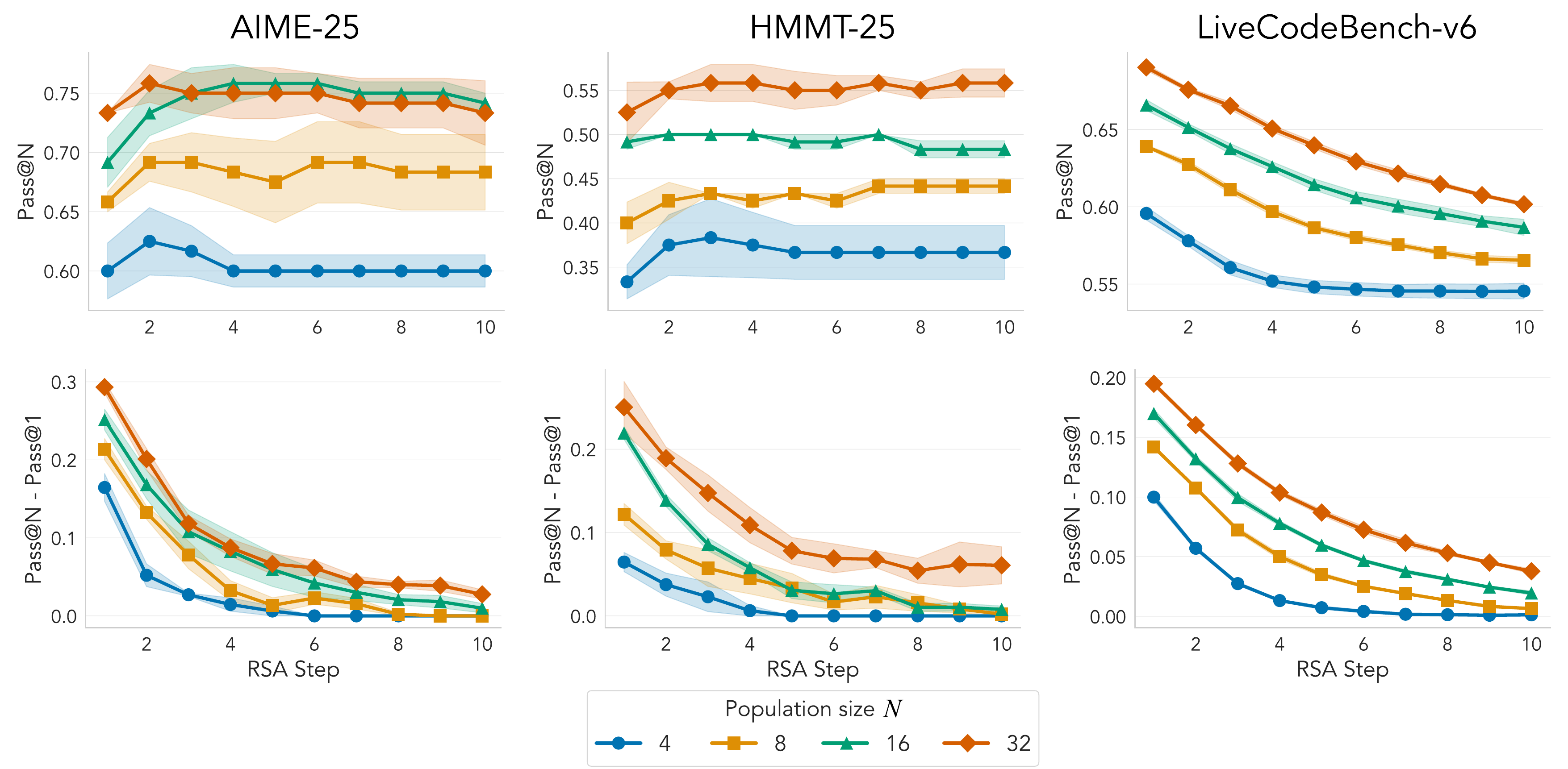}
    \caption{Pass@N (\textbf{top row}) and Pass@$N$ $-$ Pass@$1$ (\textbf{bottom row}) across \method steps for different values of $N$.  Larger $N$ results in higher Pass@N score, but requires more steps to mix, delaying the convergence of Pass@1 to Pass@$N$. All results with fixed $K=4$.}
    \label{fig:passN_N_dif}
    \vspace*{-1.5em}
\end{figure*}

\paragraph{Pass@N as a predictor of asymptotic performance.} 
\label{app:pass_N_analysis}
The Pass@$N$ score for a population of $N$ solutions is equal to $1$ if at least one final answer out of the $N$ is correct. 
The top row of \Cref{fig:passN_N_dif} shows the average Pass@$N$ score of the population across iterations of \method for different values of $N$. 
For the math tasks (AIME-25, HMMT-25), Pass@$N$ remains relatively stable, whereas for LiveCodeBench-v6 it decreases by $6$-$8\%$. As expected, larger $N$ yields a higher baseline Pass@$N$ score. 

We find that the gap between Pass@$N$ and Pass@$1$
is a useful predictor of the `aggregability' of a set of solutions. The bottom row of \Cref{fig:passN_N_dif} shows this gap over iterations. 
As the number of \method iterations  grows, Pass@$1$ converges to Pass@$N$, which acts as an upper bound on the performance. The Pass@$N$ $-$ Pass@$1$ gap consistently drops faster for smaller $N$ with fixed aggregation set size $K$. Intuitively, \method preserves good reasoning patterns in the population, and high-quality reasoning chains can mix within the population in fewer aggregation iterations if the population size is small.  
Therefore, a larger population size $N$ enables better asymptotic performance, but requires either more sequential iterations $T$ or faster mixing via larger aggregation size $K$. See \Cref{app:diversity_analysis} for a population diversity analysis over \method steps, which further validates these findings.

\begin{figure*}[t]
    \centering
    \includegraphics[width=0.95\linewidth]{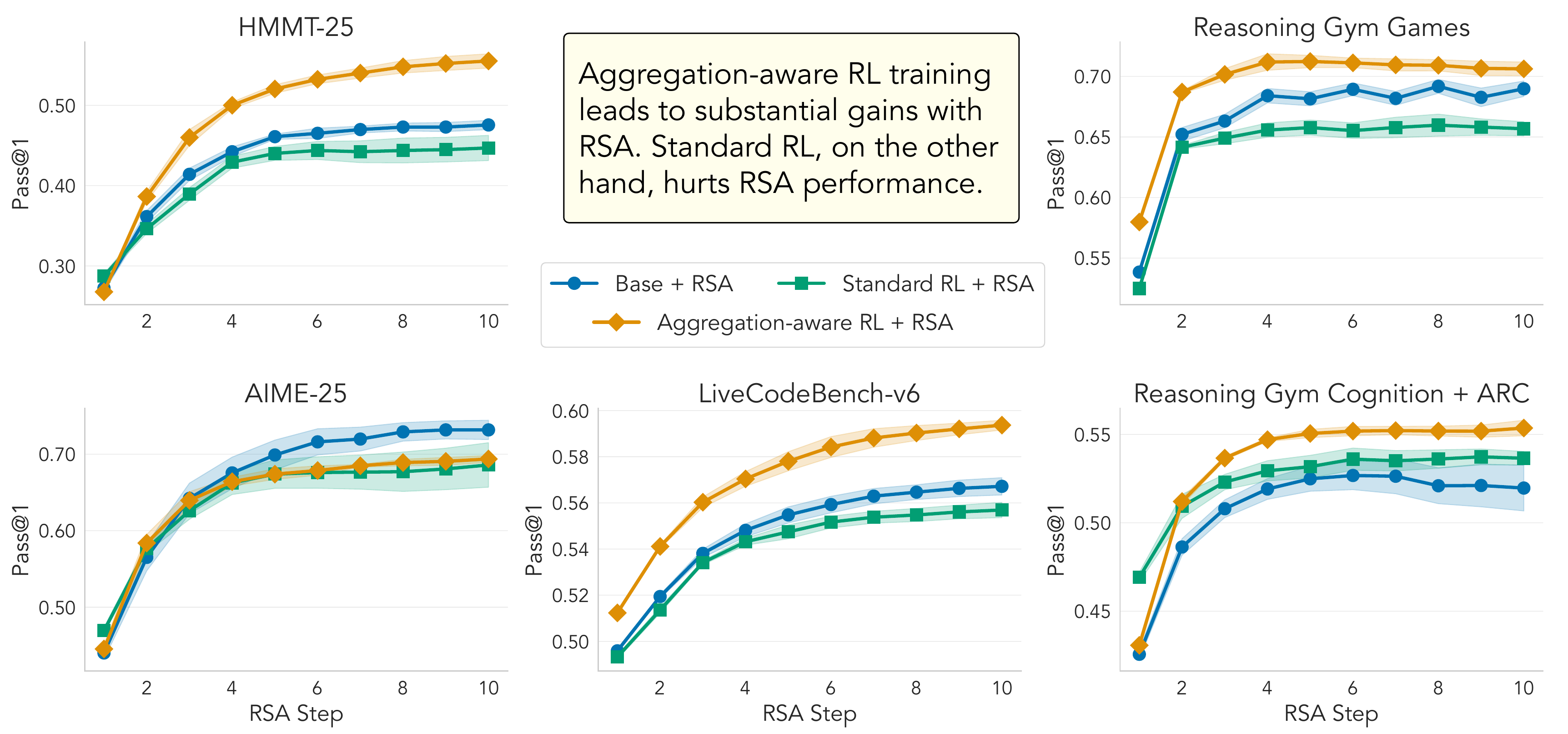}
    \caption{Pass@$1$ across \method steps for the base, standard RL, and aggregation-aware RL policies with \texttt{Qwen3-4B-Instruct-2507}. Standard RL training generally hurts performance when using \method, whereas aggregation-aware RL typically leads to improvement.}
    \label{fig:rl_results}
    \vspace*{-1em}
\end{figure*}
\paragraph{Tuning \method under compute budgets.}
Our results indicate that jointly increasing $N$, $K$, and $T$ improves \method performance. In practice, the key question is \emph{how to scale them relative to one another given a limited compute budget}. 
Based on the above analysis we note that:
\begin{itemize}[left=0cm,nosep]
\item Population size $N$ controls the asymptotic performance.
\item Larger aggregation set size $K$ for a fixed $N$ leads to faster mixing of high quality chains (for $K > 1$).
\item Longer self-aggregation depth $T$ monotonically improves performance.
\end{itemize}
When a higher number of sequential reasoning steps $T$ are feasible, it allows for a smaller $K$ provided $N$ is large. Conversely, when $T$ is limited due to time constraints and increasing $K$ is impractical (\eg, due to context length constraints), $N$ should also be reduced; a large population that fails to mix effectively is less useful than a smaller batch that evolves rapidly together. We expect these findings to generalize to other model families.

\subsection{\method Improves with Aggregation-Aware RL}
\label{sec:rl_exp}

We next analyze the impact of the aggregation-aware RL training procedure described in \Cref{sec:rl}.

\paragraph{Setup.}  We use \texttt{Qwen3-4B-Instruct-2507} as the reference model and construct a reasoning dataset containing 16,000 math problems from DeepScaleR \citep{deepscaler2025}, and 2048 problems each from six Reasoning Gym tasks where the reference model performs poorly (\texttt{tower\_of\_hanoi}, \texttt{sokoban}, \texttt{knight\_swap}, \texttt{rush\_hour}, \texttt{arc\_1d}, and \texttt{sentence\_reordering}). 
For each query, we generate four candidate solutions using the reference model, which are then used to form aggregation prompts (\Cref{fig:method}). 

We train an \textit{aggregation-aware} model on this augmented dataset by jointly optimizing \Cref{eq:standard_rl} for the standard prompts and \Cref{eq:agg_rl} for the aggregation prompts. As a baseline, we also train a model on the original dataset with only standard prompts by optimizing \Cref{eq:standard_rl}. The model trained using standard RL is only trained to generate solutions directly and is not optimized to aggregate reasoning chains. 
Both models are trained for 300 steps using RLOO~\citep{ahmadian2024basicsrevisitingreinforcestyle}. (Further details in \Cref{app:rl_abl}.)
For evaluation, we run \method for $T=10$ steps with the reference, \textit{standard RL} post-trained, and \textit{aggregation-aware RL} post-trained models on AIME-25, HMMT-25, LiveCodeBench-v6, and the Reasoning Gym Games and Cognition + ARC test sets from \Cref{sec:exp}. We ensure no data contamination between the training and test sets. We fix the aggregation size $K=4$ and population size $N=16$.

\paragraph{Results.}
\Cref{fig:rl_results} shows that in four out of five cases, \method with the standard RL fine-tuned model underperforms \method with the reference model, validating our hypothesis that distribution shifts incurred due to test-time scaling can lead to performance degradation after RL. The aggregation-aware policy, on the other hand, always outperforms standard RL and significantly outperforms the reference in four out of five tasks, with AIME-25 being the only outlier. Interestingly, we see large gains on LiveCodeBench despite our training dataset not containing coding problems, which might indicate that the aggregation skills exhibit strong out-of-domain transferability. Overall, these experiments clearly demonstrate the benefits of aggregation-aware RL. Considering the implementation simplicity and the resulting robustness gains to \method (or even to single-step self-aggregation), we strongly encourage its adoption for post-training.

\section{Conclusion}
\label{sec:discussion}

\looseness=-1
We introduce Recursive Self-Aggregation (\method), a hybrid test-time scaling framework that treats reasoning as an evolutionary process. By recursively aggregating reasoning chains, \method enables models to cross-reference and recombine information across multiple candidates, while still retaining the depth of sequential refinement. This allows RSA to generate solutions that consistently outperform single-trajectory refinement and purely parallel scaling strategies. We further show that RL fine-tuning the LLM to perform aggregation amplifies \method's benefits, yielding superior performance.

\paragraph{Future work.} In future work, \method can be composed with other test-time scaling methods to further improve performance, for example, by using self-verification to filter low-quality candidates from the population, thus introducing an explicit fitness function to the evolutionary algorithm. Another promising idea is to use multi-step reinforcement learning to train the policy for the end-to-end \method procedure, moving beyond the greedy single-step aggregation explored in this work.

\section*{Acknowledgments}
The authors thank Emiliano Penaloza for helpful comments.

The research was enabled in part by computational resources provided by the Digital Research
Alliance of Canada (\url{https://alliancecan.ca}), Mila (\url{https://mila.quebec}), NVIDIA, and the National Energy Research Scientific Computing Center, a DOE Office of Science User Facility supported by the Office of Science of the U.S. Department of Energy under Contract No. DE-AC02-05CH11231 using NERSC award NERSC DDR-ERCAP0034652.

YB acknowledges funding from National Sciences and Engineering Council of Canada (NSERC) and the Canadian Institute for Advanced Research (CIFAR). GL acknowledges support from NSERC Discovery Grant RGPIN-2018-04821, the Canada Research Chair in Neural Computations and Interfacing, and a Canada-CIFAR AI Chair. GB acknowledges funding from NSERC and CIFAR. NM acknowledges support from the CIFAR Learning in Machines and Brains program. VS was supported by a UNIQUE scholarship. MJ is supported by a FRQNT Doctoral Fellowship (\url{https://doi.org/10.69777/366694}). SM acknowledges funding from FRQNT Doctoral Fellowship (\url{https://doi.org/10.69777/372208}).

Prepared by LLNL under Contract DE-AC52-07NA27344 and supported by the LLNL-LDRD Program under Project No. 24-ERD-058. This manuscript has been authored by Lawrence Livermore National Security, LLC under Contract No. DE-AC52-07NA27344 with the U.S. Department of Energy. The United States Government retains, and the publisher, by accepting the article for publication, acknowledges that the United States Government retains a non-exclusive, paid-up, irrevocable, world-wide license to publish or reproduce the published form of this manuscript, or allow others to do so, for United States Government purposes.



\bibliography{example_paper}
\bibliographystyle{icml2026}

\newpage
\appendix
\onecolumn

\section{Additional Related Work}
\label{app:related_work}
\paragraph{Chain-of-thought aggregation.}
Several recent papers have explored self-aggregation as a parallel scaling strategy. \citet{li2025llms} study simple single-step aggregation, while \citet{wang2025learningrefineselfrefinementparallel} enhance aggregation ability through supervised fine-tuning (SFT), which requires access to a stronger teacher LLM. Concurrent to our work, \citet{zhao2025majority} trained RL policies for single-step aggregation. Our work conducts more extensive experiments across a broader suite of tasks with ablations, and further motivate aggregation-aware RL as a means to improve the performance of test-time recursive aggregation as an additional contribution. \citet{naik2023diversity} does not use self-aggregation, instead including an ``approach'' in the prompt to generate multiple diverse solutions, from which an answer is selected using majority voting. The algorithm could easily be modified to use self-aggregation as the combination strategy instead. None of these works explored the sequential scaling and evolutionary components introduced in our work. \citet{wang2024mixture} is closely related to our approach, and uses an ensemble of LLMs to generate proposals that are jointly aggregated by a stronger model in an iterative loop. In contrast, \method uses a single LLM and mixes the population by aggregating random subsets at each step while maintaining a fixed population size greater than the aggregation size to maintain population diversity, which we identify as critical factor (see \Cref{app:pass_N_analysis,app:diversity_analysis} for further analysis). 
\paragraph{Evolutionary methods.} Another line of work closely related to \method is using LLMs within evolutionary algorithms. \citet{yang2023large} propose using LLMs as proposers and mutators within an evolutionary optimization loop. They assume access to an external fitness function to evaluate the solutions. \citet{romera2024mathematical} propose FunSearch which builds upon a similar idea using LLMs to modify and propose new python functions given a scoring function. Similar to our aggregation-aware RL approach, EvoTune~\citep{surina2025algorithm} trains the LLM within an evolutionary process with RL to improve the LLM in the context of program synthesis.

\paragraph{Other related works.}
Several hybrid scaling strategies build on Tree-of-Thoughts (ToT)~\citep{yao2023tree}. \citep{inoue2025wider}, expands trees over coherent text units (“thoughts”) and applies adaptive branching to Monte Carlo Tree Search, with external or self-verification serving as the value function. Graph-of-Thoughts (GoT)~\citep{besta2024graph} generalizes Tree-of-Thoughts by organizing reasoning units (“thoughts”) into a directed acyclic graph (DAG), allowing nodes to have multiple parents (through aggregation) in addition to branching. Forest-of-Thought \citep{bi2025forest} expands multiple ToTs in parallel, and combines their final solutions using majority voting; if no consensus is found, it uses an LLM to compare the reasoning processes and outcomes of the different trees to give a final answer. A key weakness of these approaches is their reliance on either external verification of outcomes or value functions for scoring partial reasoning chains, the latter being a notoriously difficult problem. They also typically require careful prompting to ensure that the generated chains consist of meaningful atomic ``thoughts''. To date, we are not aware of applications of these methods to long CoT reasoning models.

\paragraph{Concurrent Work.} In concurrent work, \citet{wang2025dynscaling} proposed a similar hybrid scaling algorithm to \method. \citet{madaan2025rethinking} also proposed a pipeline that shares similarities to \method. A key difference from \citet{madaan2025rethinking} is that in their proposed method the prompts used to generate candidates for the next iteration all share the same context \textbf{c} that summarizes reasoning chains from the previous iteration's population. In contrast, \method generates candidates by independently sampling reasoning chains from the current population, avoiding a single shared summary context. Beyond the methodological differences, our work provides substantially more extensive empirical evaluation across a broader set of tasks and models, a comprehensive ablation study over key parameters of the approach, and an additional contribution of aggregation-aware RL, which is not explored in \citet{wang2025dynscaling}. 

\section{RSA Algorithm}
\label{app:algorithm}

\begin{algorithm2e}[H]
\caption{Recursive Self-Aggregation (\method)}
\label{alg:rsa_algo}
\DontPrintSemicolon
\SetKwInput{Input}{Input}
\SetKwInput{Output}{Output}
\Input{
LLM $p_\theta$ with fixed parameters, problem $\rvx$, population size $N$, subset size $K$, steps $T$.
}
\Output{Final population $\gP_T = \{\tau^{(T)}_{1},\dots,\tau^{(T)}_{N}\}$.}

\BlankLine
\tcp{Initialization}
$\{\tau_i^{(1)}\}_{i=1}^N \sim p_{\theta}(\,\cdot \mid \rvx)$\\
$\gP_{1} \gets \{\tau^{(1)}_{1},\dots,\tau^{(1)}_{N}\}$

\BlankLine
\tcp{Recursively for $t=1,\dots,T-1$}
\For{$t \gets 1$ \KwTo $T-1$}{
  \tcp{Subsampling}
  \For{$i \gets 1$ \KwTo $N$}{
    Sample $\{m^{(t)}_{i,1},\dots,m^{(t)}_{i,K}\} \sim \mathrm{Uniform}\left(\{1,\dots,N\}\right)$ indices without replacement\;
    Form aggregation set $S_i^{(t)} \gets \{\tau^{(t)}_{m^{(t)}_{i,1}},\dots,\tau^{(t)}_{m^{(t)}_{i,K}}\}$
  }
  $\gS_t \gets \{S_1^{(t)},\dots,S_N^{(t)}\}$

  \BlankLine
  \tcp{Self-aggregation}
  \For{$i \gets 1$ \KwTo $N$}{
    $\tau_i^{(t+1)} \sim p_{\theta}(\,\cdot \mid S_i^{(t)}, \rvx)$
  }
  $\gP_{t+1} \gets \{\tau^{(t+1)}_{1},\dots,\tau^{(t+1)}_{N}\}$
}

\BlankLine
\tcp{Termination}
\Return $\gP_T$ \textbf{and optionally} sample $\hat{\tau} \sim \mathrm{Uniform}(\gP_T)$.
\end{algorithm2e}

\section{Population Diversity Analysis}
\label{app:diversity_analysis}

\begin{figure*}[h]
    \centering
    \begin{subfigure}[b]{0.45\textwidth}
        \centering
        \includegraphics[width=\linewidth]{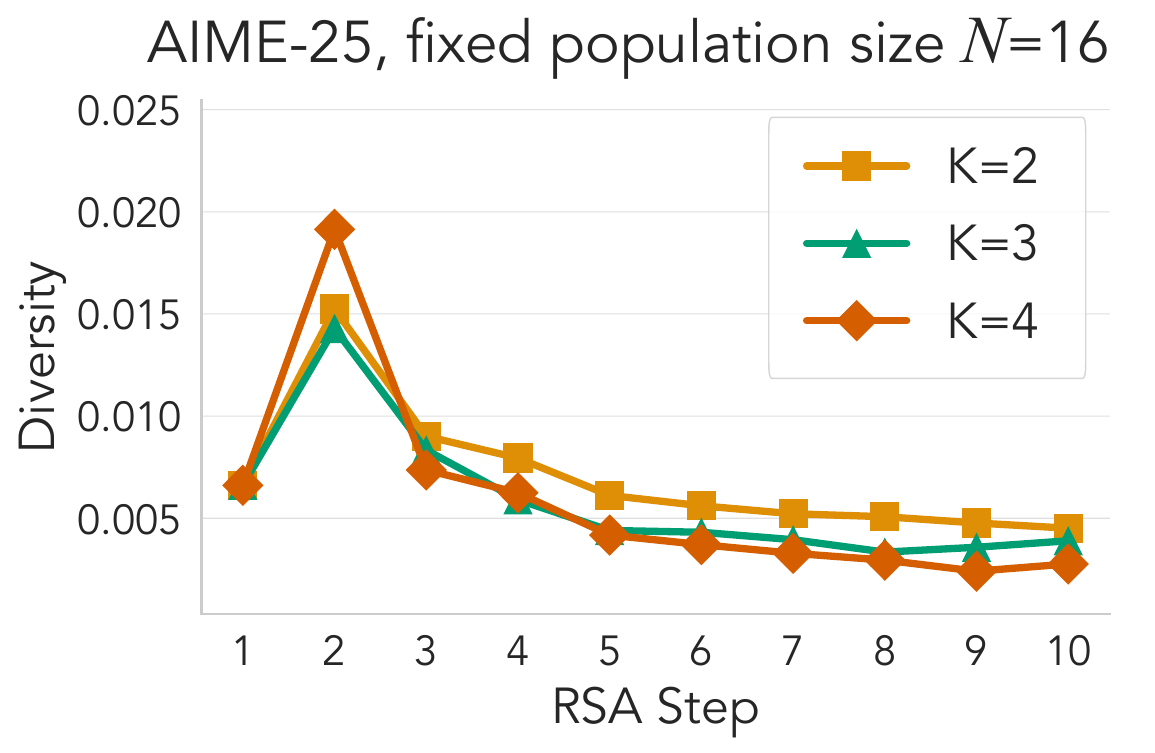}
        \label{fig:diversity_vary_K}
    \end{subfigure}
    \begin{subfigure}[b]{0.45\textwidth}
        \centering
        \includegraphics[width=\linewidth]{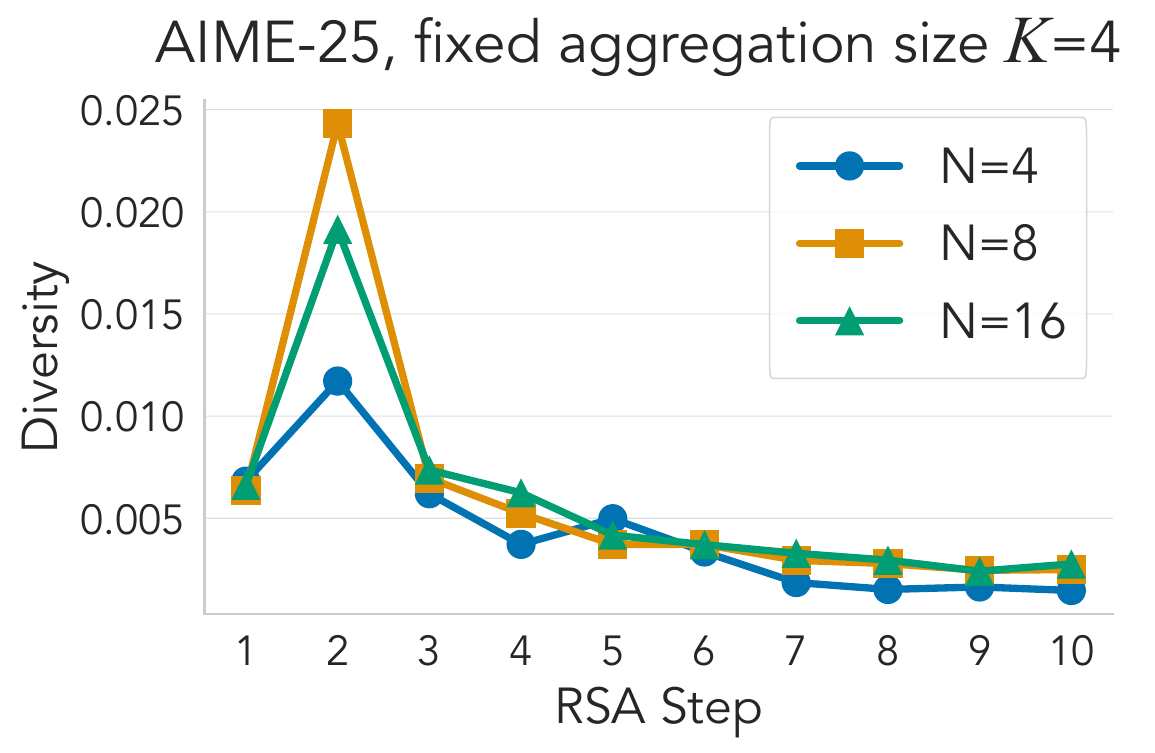}
        \label{fig:diversity_vary_N}
    \end{subfigure}
   \caption{\textbf{Left:} Diversity over \method\ steps with fixed population size $N=16$ and varying aggregation batch $K$ on AIME-25. Larger $K$ accelerates mixing, shown through faster drop in population diversity. 
\textbf{Right:} Diversity over \method steps with fixed $K=4$ and varying $N$ on AIME-25. Increasing $N$ enhances the diversity of reasoning chains, and hence the Pass@$N$ score, which determines asymptotic performance. However, very large $N$ relative to $K$ can slow mixing and hinder performance.}
\label{fig:diversity}
\end{figure*}

\paragraph{Setup.} In this experiment, we study how the diversity of the population evolves across RSA steps, using the AIME-25 dataset as an illustrative example. To quantify this, we require a metric to measure semantic diversity within a batch of CoTs. We embed the reasoning chains to generate sequence embeddings with ModernBERT~\citep{modernbert}, a strong Transformer encoder, and use the average cosine distance between embeddings in the population as a simple diversity metric. While this is an imperfect metric, it can still reveal interesting trends when plotted over time.
\paragraph{Effect of varying $K$.}
The left of \Cref{fig:diversity} plots the average population diversity across \method steps with a fixed population size $N=16$ and varying the aggregation size $K=2,3,4$. For all $K$, diversity rises sharply after the first aggregation step ($t=2$) and then steadily decays. After 10 steps, larger $K$ yields lower diversity. This aligns with our intuition and previously observed results -- larger aggregation sizes promote faster mixing of high-quality samples, leading to quicker convergence toward high-reward solutions (\Cref{fig:acc_vs_loop}). Conversely, smaller $K$ slows mixing, explaining the weaker performance observed in \Cref{fig:acc_vs_loop}.

\paragraph{Effect of varying $N$.}
The right panel of \Cref{fig:diversity} shows average population diversity across \method steps with fixed aggregation size $K=4$ and varying population sizes $N=4,8,16$. After 10 steps, diversity is lowest for $N=4$ and highest for $N=16$, though the relative differences are smaller than in the previous experiment varying $K$. Taken together with the earlier result that larger $K$ accelerates mixing and improves performance for fixed $N$, these findings support our hypothesis from \Cref{app:pass_N_analysis} -- scaling up $N$ should be accompanied by increasing $K$ or $T$, otherwise the batch will fail to mix in time and the average sample quality will remain poor after $T$ steps.

\section{Experiment Details}
\label{app:details}

In this section we provide all experiment details necessary to reproduce our results. We provide code \href{https://github.com/HyperPotatoNeo/RSA}{here}.

\subsection{Task Details} 
\label{app:tasks}
\begin{itemize}[left=0cm,nosep]
\item \textbf{Math.} We evaluate on the complete AIME-25 and HMMT-25 datasets, each consisting of 30 problems. These tasks use a binary reward: 1.0 if the predicted answer is symbolically equivalent to the ground truth and 0.0 otherwise, evaluated using \href{https://github.com/huggingface/Math-Verify}{Math-Verify}. 
\item \textbf{General reasoning.} Reasoning Gym \citep{stojanovski2025reasoninggymreasoningenvironments} consists of a broad suite of tasks divided into different categories, and difficulty levels. For our evaluations, we construct two datasets -- one from the ``Games'' category (17 tasks), requiring general reasoning and planning, and the other by combining the ``Cognition'' and ``ARC'' categories (7 + 2 tasks), requiring pattern recognition. Each of the datasets consists of 100 randomly generated problems, equally split between the tasks in the categories. We selected these categories as we found them to be the most difficult for our base models, particularly \texttt{Qwen3-4B-Instruct-2507} which we use for our ablations. We evaluate on the ``easy'' version of the problems, since the ``hard'' version is significantly more challenging, with even frontier reasoning models obtaining 0.0 reward on most tasks. The reward function is task-dependent, and can be found alongside task descriptions in the \href{https://github.com/open-thought/reasoning-gym}{Reasoning Gym repository.}
\item \textbf{Code generation.} We evaluate code generation using the complete LiveCodeBench-v6 dataset \citep{jain2024livecodebenchholisticcontaminationfree} consisting of 1055 problems. The task uses a binary reward of 1.0 if the generated Python code passes all provided test cases upon execution, else 0.0.
\item \textbf{Knowledge based reasoning.} We use SuperGPQA \citep{pteam2025supergpqascalingllmevaluation} as our knowledge-based reasoning benchmark. Although \method is designed to enhance deep reasoning, we include this dataset for completeness and still observe substantial gains. SuperGPQA consists of multiple-choice questions and assigns a binary reward of 1.0 when the selected option matches the ground truth.
\end{itemize}

\subsection{Models Used}
\label{app:models_used}
We use \texttt{Qwen3-4B-Instruct-2507} as the core model for all experiments and ablations. This choice was motivated by its small parameter count, which makes inference and RL fine-tuning tractable, while still offering strong base reasoning ability. For all models, we fix the response length to values that avoid frequent truncation and keep this constant across tasks. All model details, including their characteristics and response lengths, are tabulated in \Cref{tab:model_diversity}.

\subsection{Baseline Details}
\label{app:baselines}
\begin{itemize}[left=0cm,nosep]
    \item \textbf{Rejection sampling.} We prompt the model to self-verify $N=160$ candidate solutions. We then compute the mean score over the positively sampled solutions -- equivalent in expectation and lower variance than sampling one of these at random.
    \item \textbf{Self-refinement.} For $T=10$ steps, generated solutions are fed back into the model, which is prompted to detect errors and refine its reasoning chain.
    \item \textbf{Majority voting.} We extract the final answers from all $N=160$ reasoning chains and group equivalent ones. The majority group is then selected and compared with the ground truth.
    \item \textbf{Self-aggregation.} Equivalent to \method with a single step of aggregation, we first generate a batch of $K=4$ solutions and aggregate them with the model to produce the final answer. We stick to $K=4$ for self-aggregation since the context lengths cannot scale beyond this point without model performance degradation. As a result, this baseline is not ``budget-matched'' with \method similar to the other parallel baselines above, but we note that the ability to grow the effective batch size well above the context constraints of the model is one of the major advantages of recursive aggregation over single-step aggregation.
\end{itemize}

\subsection{RL Training Setup}
\label{app:rl_abl}
We use \texttt{verl} \citep{sheng2024hybridflow} as our framework to train the RL policies described in \Cref{sec:rl_exp}. The aggregation-aware dataset is a 50-50 split between standard and aggregation prompts. To generate the aggregation prompts, we use our standard inference procedure \Cref{app:llm-inference} with the \texttt{Qwen3-4B-Instruct-2507} reference policy to generate $K=4$ candidate reasoning chains per query. All RL training parameters are shared between the standard and aggregation-aware RL training runs. We use RLOO \citep{ahmadian2024basicsrevisitingreinforcestyle} as the training algorithm with the following hyperparameters; learning rate $=1e-6$, KL coefficient $\beta = 0.0$, batch size $=256$, training steps $= 300$, response length $= 8192$, and max prompt length $=33792$ to fit the aggregation prompts.

\subsection{LLM Inference Settings}
\label{app:llm-inference}
For consistency and fairness, we share the same inference settings across all experiments. We use vLLM \citep{kwon2023efficient} for both \method and all baselines. We keep sampling parameters consistent across experiments. We set the sampling \texttt{temperature = $1.0$, top\_p = $1.0$}, and \texttt{min\_p = $1.0$}, all of which are the default settings.

\begin{table}[]
    \centering
    \caption{List of models used in our experiments. We include Mixture-of-Experts (MoE)~\citep{shazeer2017outrageously} and state-space model (SSM) architectures~\citep{gu2022efficiently,gu2024mamba}. Our evaluation spans both non-thinking and thinking models, with appropriately larger response lengths allocated to the latter.}
    \begin{tabular}{ccccc}
    \toprule
        Model & MoE & Hybrid-SSM & Thinking & Response Length \\
        \midrule
        \texttt{Qwen3-4B-Instruct-2507} & \xmark & \xmark & \xmark & 8192 \\
        \texttt{Qwen3-30B-A3B-Instruct-2507} & \cmark &  \xmark & \xmark & 8192 \\
        \texttt{Qwen3-4B-Thinking-2507} & \xmark & \xmark & \cmark & 32768 \\
        \texttt{gpt-oss-20b (medium)} & \cmark & \xmark & \cmark & 16384 \\
        \texttt{NVIDIA-Nemotron-Nano-9B-v2} & \xmark & \cmark & \cmark & 16384\\
    \bottomrule
    \end{tabular}
    \label{tab:model_diversity}
\end{table}

\subsection{ARC-AGI-2 Experiments}
\label{app:arc}
We use \texttt{gemini-3-flash-preview} for all our experiments. We follow the default configs in the official arc-agi-benchmarking repository\footnote{https://github.com/arcprize/arc-agi-benchmarking}. The output tokens are contrained to $64000$, thinking level set to "HIGH", and automatic function calling is disabled. We slightly modify the system prompt, instructing the model to provide the summarized reasoning trace along with the final answer. This is necessary because the internal CoT is inaccessible with the Gemini API. The default prompt results in the model directly producing the answer without the reasoning summary, which is necessary for \method.

\section{Future Work}
In future work, \method can be composed with other test-time scaling methods to further improve performance, for example, by using self-verification to filter low-quality candidates from the population, thus introducing an explicit fitness function to the evolutionary algorithm. Another promising idea is to use multi-step reinforcement learning to train the policy for the end-to-end \method procedure, moving beyond the greedy single-step aggregation explored in this work.

\section{Qualitative Example}
\label{app:qual_example}
We present a qualitative example below, where we provide \texttt{Qwen3-4B-Instruct-2507} with four candidate solutions for the problem: ``Compute the sum of the positive divisors (including $1$) of $9!$ that have units digit $1$."

We highlight reasoning steps in the aggregated solution that are lifted from individual candidates, including parts that are newly added by the model and were not present in any candidate. We also provide the relevant parts from the candidate solutions that appear in the aggregated solution.
\begin{itemize}[left=0cm,nosep]
    \item \textbf{Step 1:} While all candidate solutions begin by calculating $9!$, some of them compute the products pairwise. The language in the aggregated solution most closely mirrors the \Cone{first candidate}.
    \item \textbf{Step 2:} Different candidates compute prime factors in slightly different ways. The only solution that mentions Legendre's formula explicitly is the \Ctwo{second candidate}.
    \item \textbf{Step 3:} All solutions identified that any multiple of $5$ cannot have a units digit of $1$. However, only the \Cfour{fourth candidate} correctly identified that even numbers also cannot have a units digit of $1$, which is then used in the aggregated solution. This significantly shortens the search space for valid divisors.
    \item \textbf{Step 4:} This step had the most diversity among the candidates. The first candidate exhaustively listed all possible combinations and pooled them into six different cases. While the aggregated solution considers the same divisors as candidate \# 4 (because of correctly identifying it can't be an even number), it arranges them into a table, which was not done in any individual candidate. This shows that the aggregated solution can add \NEW{new information} not present in any candidates.  
\end{itemize}

\begin{expblock}[]{Aggregated Solution}
We are asked to compute the sum of the positive divisors of $9!$ that have a \textbf{units digit of 1}.

\medskip\hrule\medskip

\Cone{\textbf{Step 1: Compute $9!$}}

\[
    \ConeM{9! = 9 \times 8 \times 7 \times 6 \times 5 \times 4 \times 3 \times 2 \times 1 = 362880}
\]

\medskip\hrule\medskip

\textbf{Step 2: Prime Factorization of $9!$}

\Ctwo{Using Legendre’s formula, the prime factorization is:}

\[
\CtwoM{\text{Exponent of 2: } \left\lfloor \frac{9}{2} \right\rfloor + \left\lfloor \frac{9}{4} \right\rfloor + \left\lfloor \frac{9}{8} \right\rfloor = 4 + 2 + 1 = 7}
\]
\[
\CtwoM{\text{Exponent of 3: } \left\lfloor \frac{9}{3} \right\rfloor + \left\lfloor \frac{9}{9} \right\rfloor = 3 + 1 = 4}
\]
\[
\CtwoM{\text{Exponent of 5: } \left\lfloor \frac{9}{5} \right\rfloor = 1 \qquad
\text{Exponent of 7: } \left\lfloor \frac{9}{7} \right\rfloor = 1}
\]

Thus,
\[
9! = 2^7 \times 3^4 \times 5^1 \times 7^1
\]

\medskip\hrule\medskip

\textbf{Step 3: Identify Divisors with Units Digit 1}

We are to find all \textbf{positive divisors $d$} of $9!$ such that $d \bmod 10 = 1$.

Note: 
\begin{itemize}
\item Any divisor divisible by \textbf{5} must end in \textbf{0 or 5}. Thus, \textbf{no divisor divisible by 5} can have units digit 1.
\item \Cfour{Any divisor divisible by \textbf{2} is even $\to$ ends in an even digit (0, 2, 4, 6, 8).}
\Cfour{ So \textbf{no even divisor} can end in 1.}
\end{itemize}

Therefore, any divisor ending in 1 must be:
\begin{itemize}
\item \Cfour{\textbf{Odd} $\to$ not divisible by 2 or 5}
\item \textbf{Not divisible by 5}
\item \Cfour{So must be a product of powers of \textbf{3 and 7 only}}
\end{itemize}

\Cfour{That is, the divisor must be of the form:}
\[
\CfourM{d = 3^b \cdot 7^c}
\]
\Cfour{where:}
\[
\CfourM{0 \le b \le 4 \quad (\text{since } 3^4 \mid 9!), \qquad c \in \{0,1\} \quad (\text{since } 7^1 \mid 9!).}
\]

\medskip\hrule\medskip

\textbf{Step 4: List All Such Divisors with Units Digit 1} \textcolor{white}{b}

\Cfour{We compute all $d = 3^b \cdot 7^c$ for $b = 0,\dots,4$, $c \in \{0,1\}$, and check which end in 1.}

\begin{center}
\NEW{%
\begin{tabular}{cccccc}
\toprule
$b$ & $3^b$ & $c$ & $d$ & Units Digit \\
\midrule
0 & $1$   & 0 & $1$   & 1 \cmark \\
0 & $1$   & 1 & $7$   & 7 \xmark \\
1 & $3$   & 0 & $3$   & 3 \xmark \\
1 & $3$   & 1 & $21$  & 1 \cmark \\
2 & $9$   & 0 & $9$   & 9 \xmark \\
2 & $9$   & 1 & $63$  & 3 \xmark \\
3 & $27$  & 0 & $27$  & 7 \xmark \\
3 & $27$  & 1 & $189$ & 9 \xmark \\
4 & $81$  & 0 & $81$  & 1 \cmark \\
4 & $81$  & 1 & $567$ & 7 \xmark \\
\bottomrule
\end{tabular}%
} %
\end{center}

Only the following values end in 1:
\[
1,\quad 21,\quad 81.
\]

All are divisors of $9!$:
\[
1 \text{: trivial divisor}, \qquad
21 = 3 \times 7 \text{: both powers present}, \qquad
81 = 3^4 \text{: } 3^4 \mid 9!.
\]

\medskip\hrule\medskip

\paragraph*{Step 5: Compute the Sum}

\[
1 + 21 + 81 = \boxed{103}
\]

\medskip\hrule\medskip

\paragraph*{Final Answer}
\[
\boxed{103}
\]

\end{expblock}

\begin{expblock}[colframe=blue!50!black]{Candidate \#1}
We are asked to compute the \textbf{sum of the positive divisors of $\mathbf{9!}$} that have \textbf{units digit 1}.

\medskip\hrule\medskip

\textbf{Step 1: Compute $\mathbf{9!}$}

$$9! = 9 \times 8 \times 7 \times 6 \times 5 \times 4 \times 3 \times 2 \times 1$$

We can compute it step by step:
\begin{itemize}
    \item $9! = 362880$
\end{itemize}

\medskip\hrule\medskip

$\vdots$

\end{expblock}

\begin{expblock}[colframe=green!50!black]{Candidate \#2}
We are asked to compute the \textbf{sum of the positive divisors of $\mathbf{9!}$} that \textbf{end in the digit 1} (\ie, have \textbf{units digit 1}).

$\vdots$

\medskip\hrule\medskip

\textbf{Step 2: Prime factorization of $\mathbf{9!}$}

We find the prime factorization of $9!$ using Legendre’s formula:

For a prime $p$, exponent in $n!$ is:

$$\sum_{k=1}^{\infty} \left\lfloor \frac{n}{p^k} \right\rfloor$$

Primes $\leq 9$: $2$, $3$, $5$, $7$

\textbf{Exponent of 2:}
$$\left\lfloor \frac{9}{2} \right\rfloor + \left\lfloor \frac{9}{4} \right\rfloor + \left\lfloor \frac{9}{8} \right\rfloor = 4 + 2 + 1 = 7$$

\textbf{Exponent of 3:}

$$\left\lfloor \frac{9}{3} \right\rfloor + \left\lfloor \frac{9}{9} \right\rfloor = 3 + 1 = 4$$

\textbf{Exponent of 5:}

$$\left\lfloor \frac{9}{5} \right\rfloor = 1$$

\textbf{Exponent of 7:}
$$\left\lfloor \frac{9}{7} \right\rfloor = 1$$

So,

$$9! = 2^7 \times 3^4 \times 5^1 \times 7^1$$

\medskip\hrule\medskip

$\vdots$

\end{expblock}

\begin{expblock}[colframe=orange!70!black]{Candidate \#4}
We are asked to compute the \textbf{sum of the positive divisors of $\mathbf{9!}$} that have \textbf{units digit 1}.

$\vdots$

\medskip\hrule\medskip

\textbf{Step 3: List all divisors of $9!$ that end in 1}

Rather than list \emph{all} divisors (which would be $(7+1)(4+1)(1+1)(1+1) = 8 \times 5 \times 2 \times 2 = 160$), we need only the ones ending in \textbf{1}.

We can \textbf{generate the divisors} in a smart way, or use modular arithmetic to filter.

But since the total number is $160$, and many divisors end in $0, 2, 4, 5, 6, 8$, we can hope that only a few end in 1.

Let’s consider the \textbf{last digit} of a divisor.

We can iterate through all divisors and check the last digit, but we can also use \textbf{modular filtering}:

We are to find all $d \mid 9!$ such that $d \equiv 1 \pmod{10}$

Note: If $d \mid 9!$, and $ d \equiv 1 \pmod{10}$, then $d$ is \textbf{odd} (since it ends in 1), and $ d \not\equiv 0 \pmod{5}$, because any multiple of 5 ends in 0 or 5, and only 5 ends in 5, but 1 is not 5.

So \textbf{d cannot be divisible by 5}.

That’s a key observation!

\textbf{Key Observation:}
\begin{itemize}
    \item Any divisor of $9!$ ending in 1 must \textbf{not} be divisible by 5.
    \item Because: If divisible by 5 → ends in 0 or 5 → can\'t end in 1.
\end{itemize}

So, we are only interested in \textbf{divisors of $9!$ that are not divisible by 5} and \textbf{end in 1}.

But even among those not divisible by 5, most end in 2, 4, 6, 8 — only some end in 1.

So actual candidate divisors: divisors of $\frac{9!}{5} = \frac{362880}{5} = 72576 $

Since we are excluding all divisors divisible by 5.

So, we now consider only \textbf{divisors of $ 2^7 \times 3^4$} (since we remove the 5 factor), and among them, those ending in \textbf{1}.

So the possible divisors (that are not divisible by 5) are all numbers of the form:

$$d = 2^a \times 3^b \quad \text{where } 0 \le a \le 7, \\
0 \le b \le 4$$

We want those such that $d \equiv 1 \pmod{10} $

Let’s go through all combinations of $ (a,b) $, compute $ d = 2^a \times 3^b $, and check if $ d \mod 10 = 1 $

$\vdots$

Wait — \textbf{important point:}
\begin{itemize}
    \item If $ a \geq 1 $, then $ d $ is even → last digit is even → cannot be 1.
    \item So \textbf{only when $ a = 0 $} can a divisor $ d $ be odd, hence end in 1.
\end{itemize}

Thus, \textbf{all divisors ending in 1 must have $ a = 0 $}.

$\vdots$

What about \textbf{7}?

Ah! We forgot about 7!

We initially focused on $ 2^a \cdot 3^b $, but we removed 5, but \textbf{we haven’t included the factor of 7}!

Wait — this is a \textbf{major mistake}.

We said ``divisors not divisible by 5" — that\'s correct.

But divisors can include **7** — since $ 7^1 $ divides $ 9! $

$\vdots$

\end{expblock}

\section{RSA Prompts}
\label{app:prompts}
We use a simple prompt template with minor task-appropriate changes, without tuning beyond some initial adjustments. Although greater gains may be achievable with careful prompt engineering, we avoided this to prevent skewing results, and instead report performance that can be reasonably expected from a straightforward implementation of \method. In fact, we look forward to future work that applies automated prompt optimization, such as GEPA \citep{agrawal2025gepareflectivepromptevolution}, to design prompts tailored to the end-to-end \method procedure, offering a cheaper alternative to the RL-based approach discussed at the end of \Cref{sec:discussion}. The prompt generation functions for math (AIME-25, HMMT-25), Reasoning Gym, and SuperGPQA tasks are given below. The prompt illustrated in \Cref{fig:method} is the essentially the same prompt without any task-specific formatting. The prompts used for LiveCodeBench are similar, but contain some additional instructions to condition on the provided starter code.

\begin{tcolorbox}[
  title=\method\ prompt generation code,
  colback=white,
  colframe=black!70,
  boxrule=0.6pt,
  arc=2mm,
  breakable
]
\begin{lstlisting}[style=editor]
def aggregate_prompt(query: str, candidate_answers: List[str], task: str) -> str:
    # Reasoning Gym
    if task == 'rg':
        problem_kind = 'problem'
        format_hint = '<answer>...</answer>'
    # SuperGPQA
    elif task == 'supergpqa':
        problem_kind = 'multiple-choice problem'
        format_hint = '\\boxed{}. Only include the correct option letter in \\boxed{}; for example \\boxed{A}'
    # Math (AIME-25, HMMT-25)
    else:
        problem_kind = 'math problem'
        format_hint = '\\boxed{}'
    
    parts = []
    # If K=1, we are doing single-trajectory refinement
    if len(candidate_answers) == 1:
        parts.append(
            f"You are given a {problem_kind} and a candidate solution. "
            "The candidate may be incomplete or contain errors. "
            "Refine this trajectory and produce an improved, higher-quality solution. "
            "If it is entirely wrong, attempt a new strategy. "
            f"End with the final result in {format_hint}.\n"
        )
    # If K>1, we are doing multi-trajectory aggregation
    else:
        parts.append(
            f"You are given a {problem_kind} and several candidate solutions. "
            "Some candidates may be incorrect or contain errors. "
            "Aggregate the useful ideas and produce a single, high-quality solution. "
            "Reason carefully; if candidates disagree, choose the correct path. "
            "If all are incorrect, then attempt a different strategy."
            f"End with the final result in {format_hint}.\n"
        )
    
    parts.append("Problem:\n")
    # Original query is appended here.
    # It also contains the formatting instructions.
    parts.append(query.strip() + "\n")
    
    # If K=1, we are doing single-trajectory refinement
    if len(candidate_answers) == 1:
        parts.append("Candidate solution (may contain mistakes):\n")
        ans_str = (candidate_answers[0] or "").strip()
        parts.append(f"---- Candidate ----\n{ans_str}\n")
        parts.append(
            f"Now refine the candidate into an improved solution. "
            "Provide clear reasoning and end with the final answer in {format_hint}."
        )
    # If K>1, we are doing multi-trajectory aggregation
    else:
        parts.append("Candidate solutions (may contain mistakes):\n")
        for i, ans in enumerate(candidate_answers, 1):
            ans_str = (ans or "").strip()
            parts.append(f"---- Solution {i} ----\n{ans_str}\n")
        parts.append(
            f"Now write a single improved solution. Provide clear reasoning and end with the final answer in {format_hint}."
        )
    
    return "\n".join(parts)

def build_prompt(query: str, candidate_answers: Optional[List[str]], task: str):
    # At t=1, candidate_answers is None and we return base query
    # At t>1, candidate_answers is a list of strings and we return an aggregated prompt
    if candidate_answers is not None:
        prompt = aggregate_prompt(query, candidate_answers, task)
    else:
        prompt = query
    return prompt
\end{lstlisting}
\end{tcolorbox}

\end{document}